\documentclass[conference]{IEEEtran}
\usepackage{times}

\usepackage[numbers]{natbib}
\usepackage{multicol}
\usepackage{graphics,graphicx,caption,float,subcaption,booktabs,xcolor,multirow,array,color,ifthen,tabu,colortbl,dblfloatfix,url,xparse,mathtools,patchcmd,algorithm,algorithmic,amssymb,xspace,nicefrac,microtype,amsmath,amsfonts,bm,ragged2e,tikz,stackengine,etoolbox,xpatch,enumerate,xstring,setspace,tabularx,makecell,changepage,cuted,titlesec,wrapfig,tcolorbox, sidecap}
\usepackage[para]{footmisc}
\usepackage[pagebackref=true,breaklinks=true,colorlinks=true,bookmarks=false,citecolor=blue]{hyperref}
\usepackage{algorithm}

\begin{document}
\newcommand{\our}{WHIRL\xspace}
\newcommand{\ours}{WHIRL\xspace}
\newcommand\marginfootnote[1]{\refstepcounter{mgncount}\marginpar{{$^\themgncount$}#1}\footnotemark}

\title{Human-to-Robot Imitation in the Wild}

\author{Shikhar Bahl$\qquad$Abhinav Gupta$^\star$ $\qquad$Deepak Pathak$^\star$ \\ Carnegie Mellon University}

\makeatletter
\let\@oldmaketitle\@maketitle% Store \@maketitle
\renewcommand{\@maketitle}{\@oldmaketitle% Update \@maketitle to insert...
  \vspace{0.075in}
  \includegraphics[width=1\linewidth,height=0.58\linewidth]{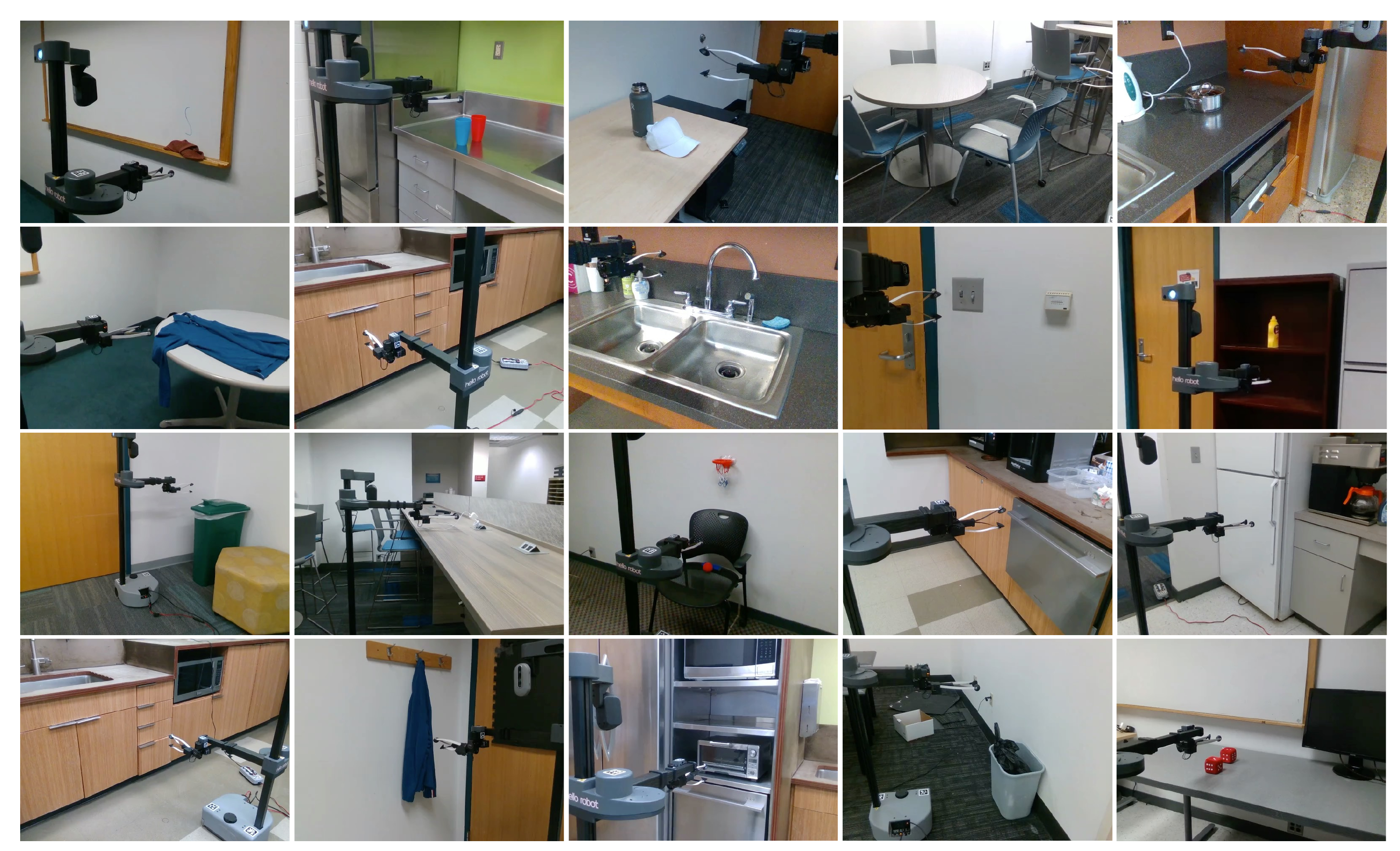}
  \centering
  \captionof{figure}{\small We present \ours, an efficient real-world algorithm for one-shot visual imitation in the wild. \ours is able to directly learn from unstructured human videos and generalize to new tasks as well. Videos and webpage at:  \texttt{\url{https://human2robot.github.io}  }
  }
  \label{fig:teaser}
  \vspace{-0.3in}
  \bigskip}
\makeatother
\maketitle
\addtocounter{figure}{-1}

\begin{abstract}
We approach the problem of learning by watching humans in the wild. While traditional approaches in Imitation and Reinforcement Learning are promising for learning in the real world, they are either sample inefficient or are constrained to lab settings. Meanwhile, there has been a lot of success in processing passive, unstructured human data. We propose tackling this problem via an efficient one-shot robot learning algorithm, centered around learning from a third person perspective. We call our method WHIRL: In-the-Wild Human Imitating Robot Learning. WHIRL extracts a \textit{prior} over the intent of the human demonstrator, using it to initialize our agent's policy. We introduce an efficient real-world policy learning scheme that improves using interactions. Our key contributions are a simple sampling-based policy optimization approach, a novel objective function for aligning human and robot videos as well as an exploration method to boost sample efficiency. We show one-shot generalization and success in real world settings, including 20 different manipulation tasks in the wild. Videos at \url{https://human2robot.github.io}.  
\end{abstract}
\IEEEpeerreviewmaketitle

\section{Introduction}

\begin{figure*}[t]
  \centering
  \includegraphics[width=\linewidth]{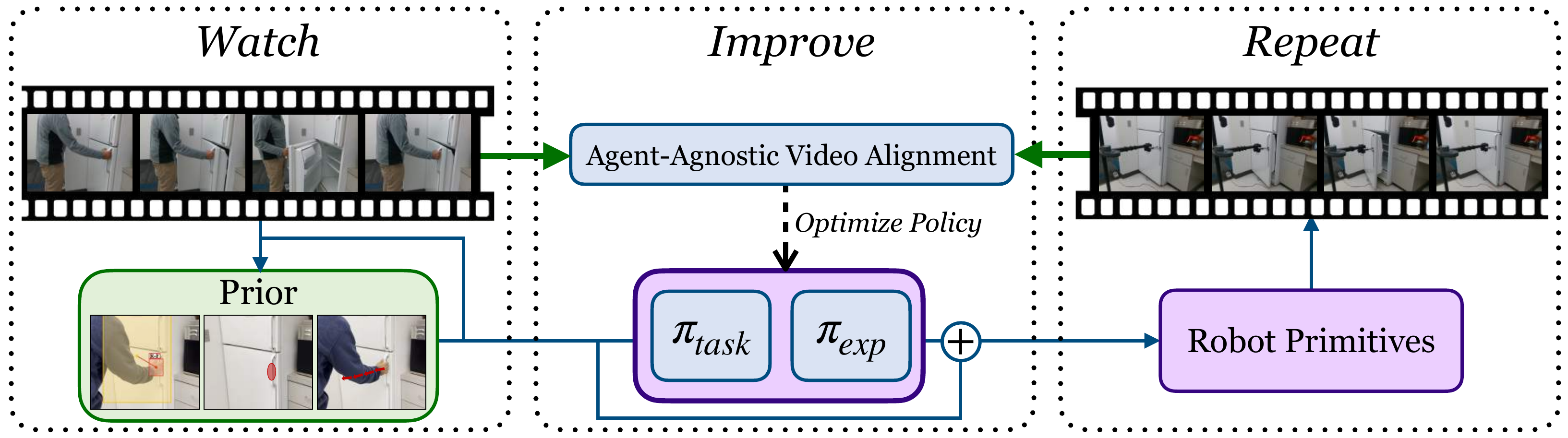}
  \caption{\small Our method (\ours) provides an efficient way to learn from human videos. We have three core components: we first "watch" and obtain human priors such as hand movement and object interactions. We "repeat" these priors by interacting in the real world, by both trying to achieve task success and explore around the prior. We "improve" our task policy by leveraging our agent-agnostic objective function which aligns human and robot videos. }
  \vspace{-0.1in}
  \label{fig:method}
\end{figure*} 

In recent years, there has been significant advances in robot manipulation: from grasping to pushing and pick/place tasks~\cite{andrychowicz2017hindsight,mahler2017dex,kalashnikov2021mt}; from manipulating a rubik's cube~\cite{akkaya2019solving} to opening cabinet doors or makeshift doors~\cite{sermanet2016unsupervised, pong2019skew}. While there has been substantial progress, most experiments in this area have still been restricted to simulation~\cite{makoviychuk2021isaac, openaigym, todorov12mujoco} or table-top experiments in the lab~\cite{levine2016learning, dasari2021rb2}. We ask a basic question as to why hasn't this progress transferred to manipulation in the real-world setup and why do we still see most experiments in lab setups or simulations? Although there have been efforts to perform grasping in home setups~\cite{gupta2018robot,song2020grasping}, general manipulation is still studied in either simulation or lab-like settings.This work delves the question of how we could move from from lab experiments to more in-the-wild setups.

We believe the biggest bottleneck for learning manipulation \textit{in the wild} is the lack of scalable and safe frameworks. Traditionally, designing a controller or policy for manipulation tasks requires learning via reinforcement (RL), which can be data-hungry and unsafe especially in the real world. While RL has had success in simulated tasks, real world tasks do not have structured rewards, thus making the problem that of sparse search. A popular alternative is to use imitation learning (IL) based approaches, but common IL approaches rely on lots of kinesthetic or teleoperated demonstrations per task. However, this data can be expensive to obtain in the real world and may not be generalizable to new settings or different robots. There have been attempts at one shot imitation at inference, but these methods requires thousands of demonstrations or interactions during training~\cite{duan2017one,finn2017one,pathak2018zero}.

To move towards general robot manipulation out of the lab and tabletop settings, we believe visually imitating humans provide a safe and scalable alternative. Rather than asking humans to teleoperate, robot should observe humans to learn as they interact in the world. Humans provide a rich source of data as they often act in interesting and near optimal ways given a task or an environment. In the visual imitation framework, the agent observes other agents perform action without access to actions (just the pixels). This data is then used to guide their own exploration and learning. However, there are several challenges in making visual imitation learning work: first, there is the issue of embodiment mismatch (robots and humans are different agents and have different bodies). Second, there is no access to actions from humans, which have to be inferred. Third, there is no access to task information including rewards beyond pixels. Current approaches use end-to-end learning~\cite{sharma2019third,yu2018one,smith2019avid} which requires a lot of samples during training and are hence restricted to lab/simulation settings.

In this paper, we propose to revive this visual human-imitation framework to move robot manipulation out of the lab and into the wild. Instead of learning end-to-end from scratch, we propose leveraging advances in computer vision and computational photography to (a) infer a trajectory and interaction information from the human, thus obtaining a prior; (b) learning an improvement policy via interactions in the real world; and (c) bridging the embodiment gap between human demonstrations and robot videos. But above all, we only use imitation data as \textbf{priors} for our policy. Naively using priors will not result in success due to a host of issues, e.g.~varying morphologies or inaccuracies in detections. It is crucial to \textit{interact} with the real world to learn generalizable manipulation. We introduce a sampling based optimization framework, similar to the Cross Entropy Method (CEM), in order to iteratively improve the interaction policy. To make \ours operate without supervision, we introduce an \textit{agent-agnostic} alignment objective function for the described optimization approach. In order to not be too restricted by the prior, we employ a novel task-agnostic exploration policy which allows the agent to sample new and interesting actions. This all leads to an efficient framework for manipulation tasks in real world.

We demonstrate our framework on 20 different tasks in 3 different environments. We show one-shot, in the wild generalization and success in various real world settings, including manipulation tasks such as opening and closing doors or fridges, putting objects in shelves, folding shirts, cleaning white boards, opening taps and a variety of other tasks. We analyze our approach thoroughly in terms of task success, generalization, and performance compared to state-of-the-art baselines. To the best of our knowledge, this is the first effort that takes robot manipulation out of the lab and into the real world at this scale.

\section{Related Work}
\subsection{Detecting Humans}

The field of computer vision has studied the problem of detecting humans in a wide variety of approaches. Most such applications are contained in the domain of graphics, but many have applications in real world robotics as well. There are many possible uses of such, for example modeling human bodies, detecting poses, inferring dynamics or understanding interactions between humans and the world. From a modeling perspective, works such as MANO \cite{MANO:SIGGRAPHASIA:2017} and SMPL \cite{loper2015smpl} have proposed analytical models of human hands and bodies respectively. Hand and body pose estimation \cite{wang2020rgb2hands, hmr, FrankMocap_2021_ICCV} can be useful in the context of robotics, as it can allow for spatially grounding a demonstration, which is something that we leverage in our approach. Estimation and detection of humans, while useful, does not help in understanding \textit{what} the human is doing. For this, large annotated video datasets can help detect and infer human actions, such as the Something-Something \cite{goyal2017something}, YouCook \cite{das2013thousand}, ActivityNet datasets \cite{caba2015activitynet} or the 100 Days of Hands \cite{100doh} (100DOH) dataset.  100DOH \cite{100doh} is particularly useful as it contains object level interaction annotations. \ours aims to remain as general as possible in terms of the human prior used, using only object interaction data. We employ both the models from \citet{FrankMocap_2021_ICCV} and the hand-object detector from \citet{100doh} for estimating hand position and interaction information. It is possible to combine \ours with stronger priors for human hands, for example, building a knowledge graph of objects and functional grasps \cite{murali2020same} or using heat sensing \cite{Brahmbhatt_2019_CVPR} to understand interactions.

\subsection{Imitation and Reinforcement Learning from Videos}

\noindent\textbf{Learning From Human Videos$\quad$}
A large field of robot learning (Learning from Demonstrations: LfD) is focused on learning from expert demonstrators \cite{pastor2009motorskills, ratliff2007imitation, pomerleau1989alvinn, ross2011reduction}. However, most of the work in this area tackles the problem of learning from demonstrations that humans provide directly to the robot via kinesthetic teaching or teleoperation. This is an expensive way to gather data for teaching robots. On the other hands, videos of humans perfoming daily activities are widely available on the internet and can provide good semantic supervision for robotics tasks. However, extracting the right knowledge, for example aligning human videos with robot videos, is challenging. One solution is to learn a direct correspondence. The use of paired human and robot data \cite{sharma2019third, sharma2018multiple, liu2017imitation} is a common approach in this line of work. For example, \citet{sharma2019third} aim to learn to produce subgoals in the robot's perspective, conditioned on a human video. \citet{liu2017imitation} seeks to learn a translation model based on the paired demonstrations directly. Collecting paired demonstrations is challenging, and only a limited amount of data can be collected. Thus, previous work \cite{smith2019avid, zakka2022xirl} has employed cycle-consistency \cite{zhu2017unpaired, dwibedi2019temporal} to learn an unsupervised pairing. Similarly, \citet{sermanet2017time} uses a contrastive loss between frames close to and far away from the anchor point in the video, in order to obtain a representation. \citet{sermanet2016unsupervised} trains a classifier using human demonstrations, which is then used to build a reward function. Unsupervised methods can learn a translation model for single tasks, however they have to be trained in every new setting, which is time consuming. Most such approaches require many random interactions to learn representations, and this process often yields unstable models \cite{smith2019avid}. \ours, on the other hand, does not need any random data to learn representations, and can work with even a single demonstration, in a variety of in-the-wild settings.

\vspace{2mm}\noindent\textbf{Offline Videos and Datasets$\quad$}
Instead of using human videos, recent approaches have attempted to employ a reacher-grabber tool as the demonstration collection device \cite{song2020grasping, young2020visual, pari2021surprising}. These approaches have the advantage of having a smaller domain gap between robot and human actions, since the videos are in first person view. However, such a setup limits the number of tasks that are achievable, and adds considerable effort in collecting the data, since the approaches are not able to use large-scale human datasets, for example Youtube videos. On the other hand, advances in many computer vision tasks such as action recognition \cite{zhang2013actemes, goyal2017something, hara2018can, carreira2017quo, wang2018non, feichtenhofer2019slowfast}, video understanding \cite{feichtenhofer2019slowfast, lin2019tsm, liu2021video} or self-supervised representation learning \cite{he2020momentum, pan2021videomoco, chen2020simple, oord2018representation} have leveraged videos collected offline. These video datasets include the Something-Something \cite{goyal2017something}, Epic Kitchens \cite{EPICKITCHENS} or ActivityNet datasets \cite{caba2015activitynet}. These can provide important semantic information as well a high amount of visual and task diversity, which can aid in generalization.  Similarly, works such as \citet{chen2021dvd} and \citet{shao2021concept2robot} find that using a large-scale human datasets, augmented with a few demonstrations from the robot as well as task labels, can help learn a semantic action classifier which generalizes to new tasks. Unlike these approaches, we do not use any task labels or robot specific fine-tuning for the feedback module. Embedding task specific knowledge into reward classifiers does not scale to in-the-wild settings, contrary to our approach.

\vspace{2mm}\noindent\textbf{Learning Action Policies from Priors$\quad$}
While learning reward functions and representations from offline videos can be useful in robotics, videos of humans contain stronger priors.  Learning keypoints \cite{chen2021unsupervised, kulkarni2019unsupervised, xiong2021learning} or object-level \cite{pirk2019online, scalise2019improving} from videos, and using these as input to a control policy  has been shown to be useful for certain tasks, but requires knowledge of the task and careful design, for example knowing the number of objects or keypoints. This can be a limiting factor when trying to scale to a general robot setup. Previous approaches have also used hand \cite{lee2017learning, nguyen2018translating} and object tracking \cite{yang2015robot} to learn action policies, however, these have been limited to simple settings and require very structured planning algorithms that are task specific. Our approach on the other hand is flexible and works for almost any manipulation task. Previous approaches do not perform any iterative improvement, contrary to \ours.

\begin{figure}[t]
\centering

\begin{subfigure}[b]{0.32\linewidth}
    \includegraphics[width=\linewidth]{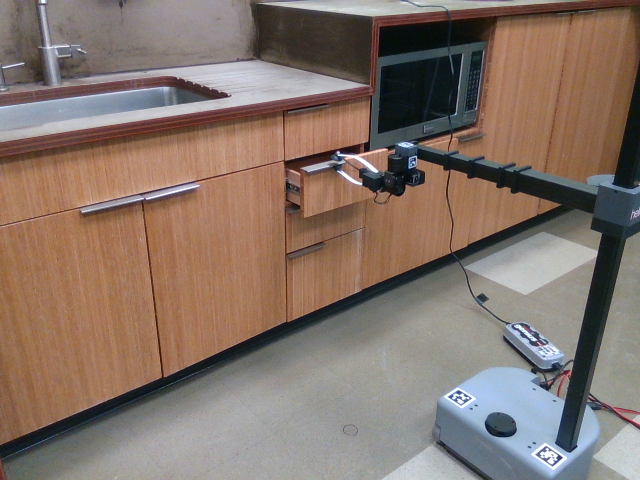}
    \vspace{-0.13in}
    \caption{\small Drawer}
    \label{fig:env-dw}
\end{subfigure}
\begin{subfigure}[b]{0.32\linewidth}
    \includegraphics[width=\linewidth]{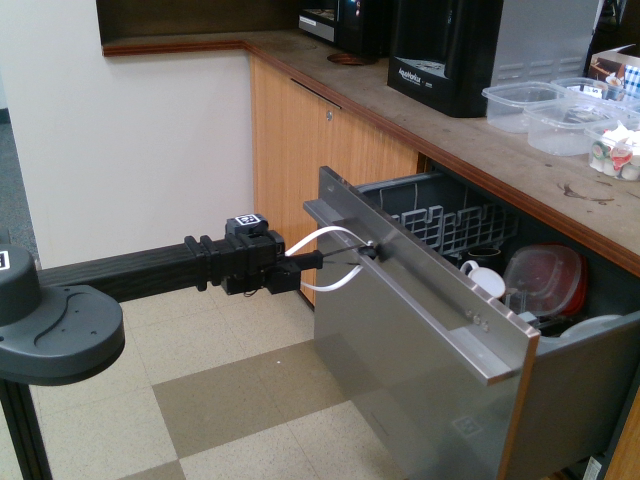}
    \vspace{-0.13in}
    \caption{\small Dishwasher}
    \label{fig:env-dr}
\end{subfigure}
\begin{subfigure}[b]{0.32\linewidth}
    \includegraphics[width=\linewidth]{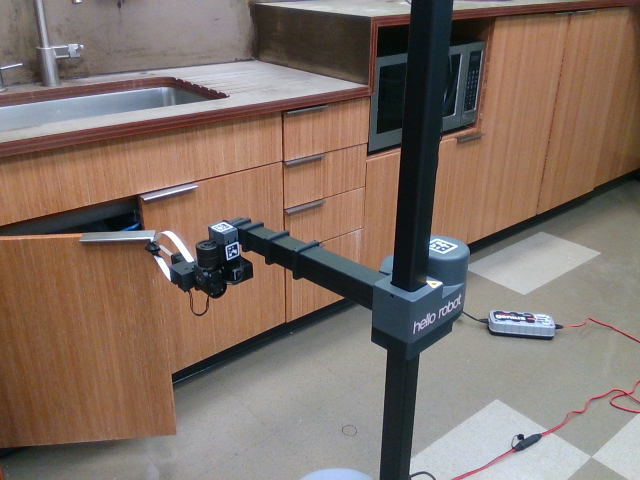}
    \vspace{-0.13in}
    \caption{\small Door}
    \label{fig:env-ds}
\end{subfigure}
\vspace{-0.04in}
\caption{\small We perform various experiments in the wild. We select a subsample of tasks, as shown above, to perform a thorough study of our \ours as well as baselines and ablations. These tasks are: drawer, door and dishwasher opening and closing.}
\vspace{-0.1in}
\label{fig:setup}
\end{figure}

\section{Human-to-Robot Visual Imitation In the Wild}

We address the challenge of learning from humans by extracting priors from observing their actions, leveraging the priors to learn an interaction policy in the real world, and exploring around the prior in an efficient manner. We build a general robot learning algorithm that can work in many in-the-wild settings. We call this approach \ours: In-the-Wild Human Imitating Robot Learning. In this section, we describe how \ours works.

\subsection{Human Priors}

\subsubsection{Extracting Human Priors}

Most trajectories ($\tau$) of interest for manipulation tasks can be broken down into smaller sub-trajectories: $\tau_{\text{pre-interaction}}$, $\tau_{\text{interaction}}$ and  $\tau_{\text{post-interaction}}$. Throughout the paper, we refer to these as primitives.  A more complex task can be thought of as a composition of such primitives. Once we are able to use human videos to estimate these primitives, we can try to deploy these on a robot, despite any differences in morphology. Videos of the desired task ($V$), such as door opening, are used to obtain this trajectory parameterization. The key components of a video of a human performing a task include how the target is moving as well as where and when the interactions happen. We describe how we infer this information from third person videos below.

\vspace{2mm}\noindent\textbf{Extracting Hand Information $\quad$}
We process each individual frame $V_t$ of the video ($V$) at timestep $t$ to obtain an estimate of the position of the hand: $x_t, y_t, z_t$. We obtain this pose using the 100DOH detection model \cite{100doh}, built on top of Faster-RCNN \cite{ren2015faster} and trained to output hand bounding box ($b_t$). This is a continuous vector of coordinates in image space. The hand position (in the camera frame) is referred to as $h_t$. In order for the robot to grasp and interact with an object, the orientation of the wrist and the force applied on the gripper are important as well. We use the MANO \cite{MANO:SIGGRAPHASIA:2017} parameterization of hands in order to obtain these. Specifically, we use the part of the parameterization that describes the rotation of the wrist, $\theta_\text{hand}^{(t)}$.

\vspace{2mm}\noindent\textbf{Extracting Interaction Information $\quad$}
Inferring the position of the hand can give useful information, but we also need to understand when the hand interacts with an object. Detecting contact is important in determining $\tau_{pre-interaction}$: it determines where the interaction occurs. Thus, we employ the 100DOH \cite{100doh} model to detect when this interaction occurs. We use this information and previously computed hand poses to extract waypoints for the robot. Specifically, we use the 100DOH model to obtain a discrete valued contact variable: $c_t$. This represents a possible contact that might be occurring at frame $t$ of the video. The possible options are: no contact, contact with portable or fixed object, and self contact. However, since out-of-the-box detections in unstructured settings can be noisy, we employ the Savitzky–Golay \cite{savitzky1964smoothing} filter for smoothing $c_t$ across timesteps. Using smoothed detection $\hat{c}_t$ we determine the time-step where the interaction started in the video: $t_{\text{interaction}}$ and when it ended: $t_{\text{end}}$. We denote the hand position at these timesteps as $h_\text{interaction}$ and $h_\text{end}$. In order to not overfit to the detections, we in fact sample from a distribution centered around the start and end points. We also sample intermediate trajectory waypoints, $h_\text{mid}$. We additionally use a simple binary representation of a grasp, determined from the contact variable $\hat{c}_t$. 

\begin{figure}[t]
\centering

\begin{subfigure}[b]{0.32\linewidth}
    \includegraphics[width=\linewidth]{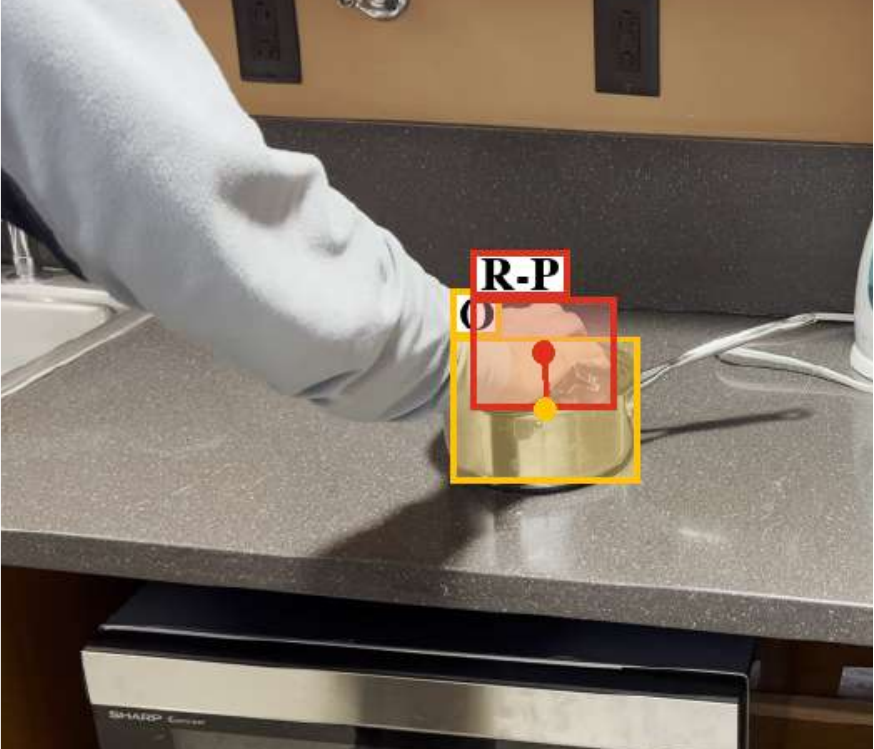}
    \vspace{-0.13in}
    \caption{\small Detection}
    \label{fig:prior-det}
\end{subfigure}
\begin{subfigure}[b]{0.32\linewidth}
    \includegraphics[width=\linewidth]{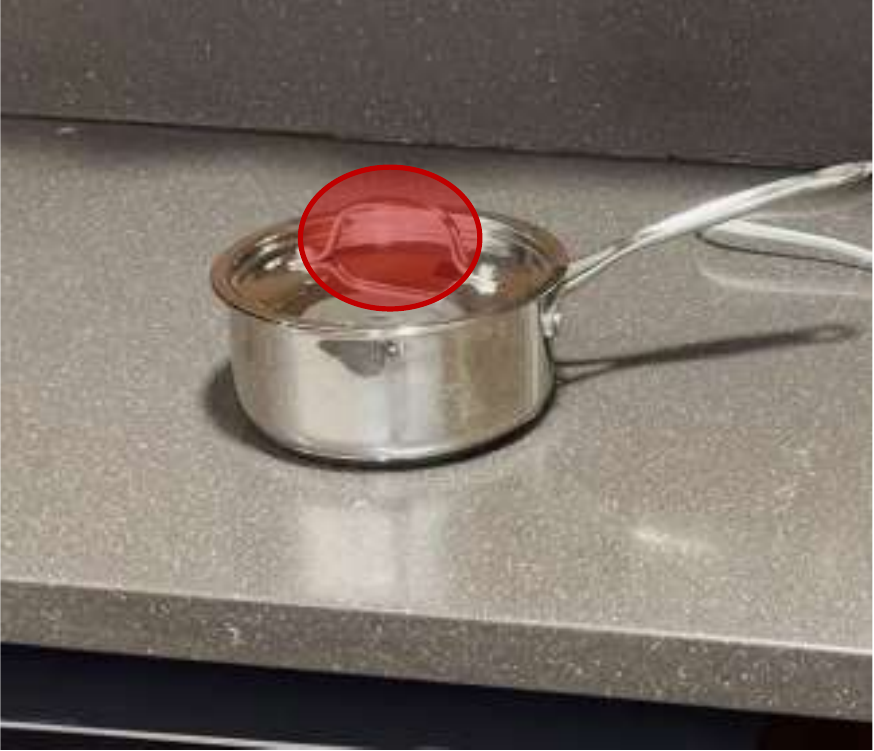}
    \vspace{-0.13in}
    \caption{\small Interaction}
    \label{fig:prior-int}
\end{subfigure}
\begin{subfigure}[b]{0.32\linewidth}
    \includegraphics[width=\linewidth]{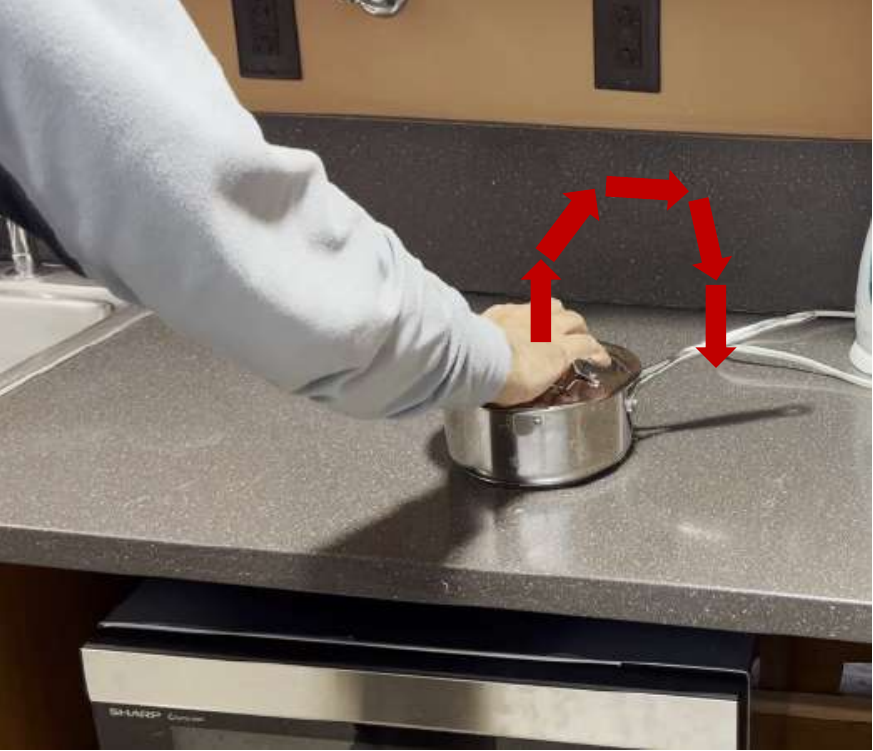}
    \vspace{-0.13in}
    \caption{\small Direction}
    \label{fig:prior-dir}
\end{subfigure}

\vspace{-0.04in}
\caption{\small We show the different components of the human prior. First we extract the position of the hand and possible object interactions (a). This indicates a possible area of interaction (b) and direction for moving the robot hand (c). We project these to the robot's action space and execute the trajectory. }
\vspace{-0.15in}
\label{fig:prior}
\end{figure}

Overall, our extracted prior from a video demonstration from a human can be described as a set of interaction waypoints: $h_\text{interaction}$, $h_\text{mid}$, and $h_\text{end}$, a grasp or interaction orientation measure $\theta_\text{hand}$, and commands to close or open the hand: $o_{1:T}$ (where $T$ is the length of the video). Figure~\ref{fig:prior} shows the different parts of the human prior we use. Note that some tasks may require a more densely sampled set of waypoints. For simplicity we think of $h_\text{mid}$ as a single point, but it can be also a set of midpoints in the hand trajectory.

\subsubsection{Converting Human Priors to Robot Priors} 

Once we obtain the desired trajectories from human videos, we can convert them into the robots frame and obtain desired poses, using depth image ($d_t$) from the external camera.  Our setup uses a depth image, but is compatible with any 3D pose estimation approach. Given a video $V_k$ of length $T$, we then project the obtained priors, $h_\text{interaction}$, $h_\text{mid}$, $h_\text{end}$, $\theta_\text{hand}$, $o_{1:T}$ to the robot's frame via 3D pose estimation from depth data. For both the gripper open and wrist orientation parameters, we use a robot-specific heuristic function, as every robot's coordinate axis is different. Let the detected waypoints be $\bm{h} = h_\text{interaction}, h_\text{mid}, h_\text{end}$. This process can be described as:
\begin{align}
f_\text{map}(\bm{h}, \theta_\text{hand}, o_{1:T}) = w_\text{interaction}, w_\text{mid}, w_\text{end}, \theta_\text{YPR}, g_{1:T} \triangleq \Psi_k 
\nonumber
\end{align}
where $\boldsymbol{w}$ are waypoints in the robot's frame, $\theta_{\text{YPR}}$ is a wrist rotation (yaw-pitch-roll format) in the robots frame, and $g_{1:T}$ are robot gripper open/close continuous parameters. We refer to this vector by $\Psi_k$.

\subsection{Policy Learning via Interaction}

Human priors from videos can give a rough guideline on how to perform the task. They are useful because they can be distilled into a neural network policy, which can possibly generalize beyond the training data. However, directly executing the prior on the task will not generally lead to success, due to differences in morphologies between human and robot hands, inaccuracies in detections, or errors in the calibration process. Thus, we need to learn a policy via real world \textit{interaction} in order to succeed at this task. Such a learning procedure must have 3 important properties: 

\begin{itemize}
    \item The real world interactions must be safe. 
    \item While safe, the interactions must not be too restrictive. 
    \item This process must be sample efficient. 
\end{itemize}

The safety of the interactions can be ensured by the human prior. Following the prior, even one that has errors, will lead to somewhat reasonable behavior, and is very likely to be safe. However, being too close to the prior will restrict the reach of the policy, and thus it will be unable able to solve the task. In order to address this challenge, we employ a \textit{task policy} which aims to solve the task and a task agnostic \textit{exploration policy} that explores around the human prior so that we do not fall into a local minimum. We describe the objective functions of these policies in the following sections. Finally, in order to ensure the learning process is sample efficient, we introduce a simple and easy to use zeroth order real world optimization procedure (similar to CEM). Since our goal is to efficiently perform many manipulation tasks in the wild, traditional RL methods are infeasible. A summary is in Algorithm 1.

\begin{algorithm}
\caption{Training Procedure for \ours}\label{alg:cap}
\begin{algorithmic}
\REQUIRE Task videos: $V_{1:K}$, $f_\text{map}$: video to robot actions function, prior task and exploration policies: $\pi$, $\pi_{\text{exp}}$. Video-level ($\Phi$) and frame-level ($\Phi_f$) agent agnostic representation. $M$ real world interactions per task. 
\WHILE{not converged}
\FOR{$k = 1...K$}
    \STATE  $\Psi_k = f_\text{map}(V_k)$ 
    \FOR{$m = 1...M$}
        \STATE  Sample $\Delta \Psi_{k, m} = \pi_{\text{exp}}(V_k, \Psi_k)$ (prob: $p$)
        \STATE  Sample $\Delta \Psi_{k, m} = \pi(V_k, \Psi_k)$ (prob: $1-p$)
        \STATE  $a_{j, m} = \Psi_k + \Delta \Psi_{k, m}$
        \STATE Execute $a_{k, m}$, collect video: $R_{k, m}$
    \ENDFOR
\ENDFOR
    
\FOR{$j = 1...K$}
    \STATE rank \texttt{Cost}$(\Phi(R_{k,m}), \Phi(V_k))$ for every $m$
    \STATE pick $E$ = \{elite examples\}
    \STATE fit $\pi(.)$ as a VAE to $\Psi_{k, m} \in E$
    \STATE pick $E_\text{exp}$ = \{$\Phi_f(R_{k,m})$ with highest "change"\}
    \STATE fit $\pi_\text{exp}(.)$ as a VAE to $\Psi_{k, m} \in E_\text{exp}$ 
\ENDFOR
\ENDWHILE
\end{algorithmic}
\end{algorithm}

\subsubsection{Policy Structure}

When trying to achieve the desired task via interaction, it is easy to simply get stuck in the local minimum around the prior. Thus we not only need to train a task policy, but an exploration policy as well.

\vspace{2mm}\noindent\textbf{Task policy$\quad$}
We would like to learn an interaction policy which will allow the agent to achieve the task. Given prior $\Psi_k$ extracted from video $V_k$, we propose learning a (task and exploration) policy $\pi(\Psi_k, V_k) = \Delta \Psi_k$, outputting the residual to the prior. Residual learning is common in robot learning \cite{johannink2019residual} as it allow for the policy to search around the prior in order to avoid unsafe behavior. Using this residual structure also allows the policy to initialize close to the human prior, and then explore from the prior as a starting point. Both of our policies are neural network based. The task policy $\pi$ will try to maximize the robot's performance with respect to the demonstration video. 

\begin{figure*}[t!]
\centering
\begin{subfigure}[b]{0.24\linewidth}
    \includegraphics[width=\linewidth]{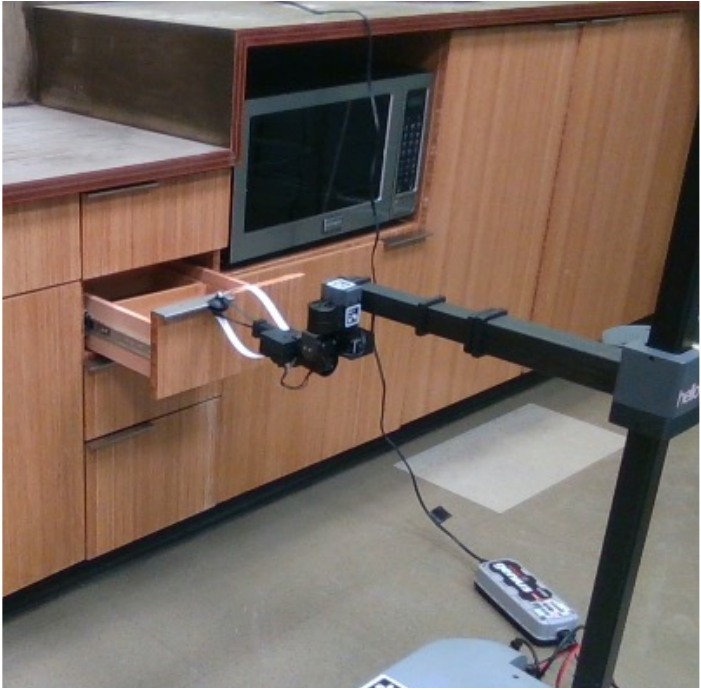}
    \vspace{-0.13in}
    \caption{\small Robot Image}
    \label{fig:im-rob}
\end{subfigure}
\begin{subfigure}[b]{0.24\linewidth}
    \includegraphics[width=\linewidth]{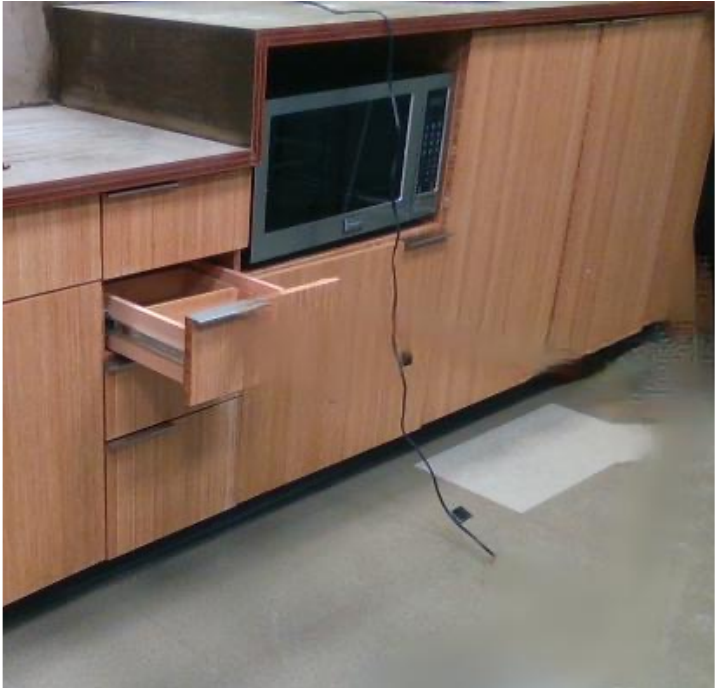}
    \vspace{-0.13in}
    \caption{\small Robot Inpainted}
    \label{fig:im-rob-inp}
\end{subfigure}
\begin{subfigure}[b]{0.24\linewidth}
    \includegraphics[width=\linewidth]{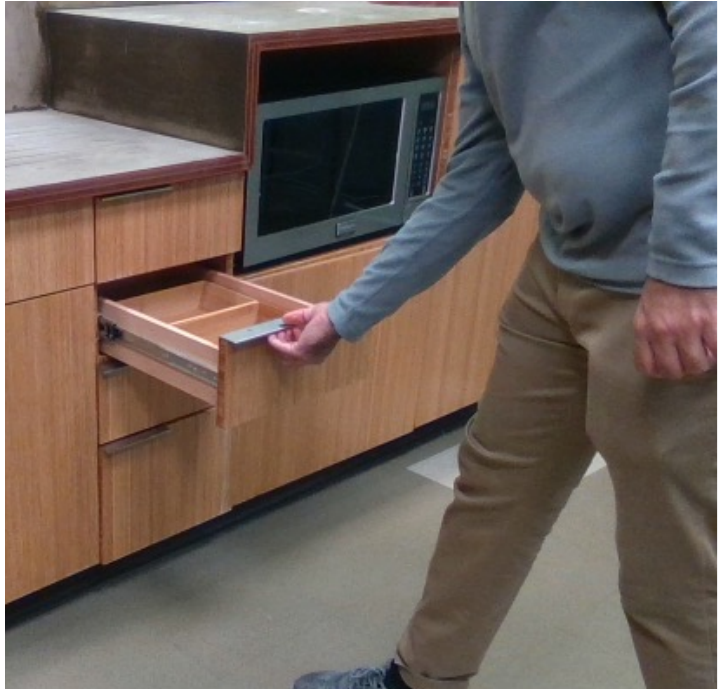}
    \vspace{-0.13in}
    \caption{\small Human Image}
    \label{fig:im-hum}
\end{subfigure}
\begin{subfigure}[b]{0.24\linewidth}
    \includegraphics[width=\linewidth]{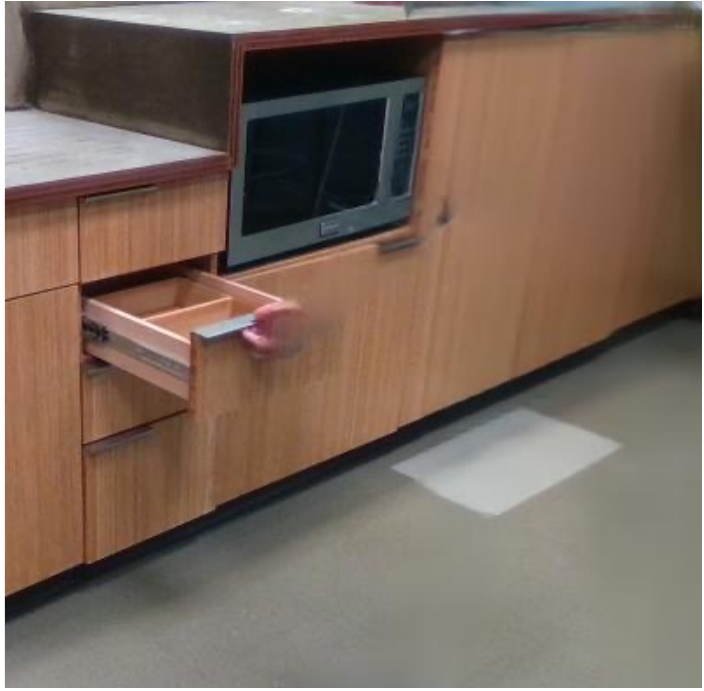}
    \vspace{-0.13in}
    \caption{\small Human Inpainted image}
    \label{fig:im-hum-inp}
\end{subfigure}
\vspace{-0.04in}
\caption{\small We show a sample of our agent agnostic representation, using the video inpainting method by \citet{lee2019copy}. An image of a human performing the task is shown in (c) and (d) shows the human inpainted from the image. Similarly for the robot, (a) and (b) show the original and inpainted images. We train segmentation models \cite{wu2019detectron2, ren2015faster} to obtain human and robot masks. }
\vspace{-0.1in}
\label{fig:inpainting}
\end{figure*}

To be able to sample around the prior, the policy needs to learn a distribution and not just a mean prediction. For a single human video, there are multiple different ways a robot can perform the task. A naive stochastic neural network policy would not be able capture this multi-modal distribution and hence has difficulty in generalizing to new videos. We leverage Variational Auto-Encoders (VAEs) \cite{kingma2013auto, rezende2014stochastic} which are popularly used to capture multi-modal distributions. In particular, we fit a Conditional VAE \cite{sohn2015learning} to learn a mapping from a set of samples $\Delta \Psi_{k, m}$ and an embedding of the input demonstration video $\phi(V_k)$. We condition the distribution on the video, $V_k$ and learn to encode the prior residuals. The encoder, $q(z |c, x)$ takes input $x = \Delta \Psi_{k, m}$, and $c$ is an embedding of the human video $\phi(V_k)$. The decoder $p(x | z, c)$ takes a latent sample from $p(z)$ as well as the human video embedding. At inference time, we can use the policy, $\pi$ to output $\Delta \hat{\Psi}_{k}$ (the residual), conditioned on video $\phi(V_k)$ and latent $z \sim \mathcal{N}(0, 1)$, where $\pi = p(x | z, c)$. Since this policy is conditioned on an input video, with enough collected data it is able to generalize to new human demonstrations as well.

\vspace{2mm}\noindent\textbf{Exploration Policy$\quad$}
On the other hand, the exploration policy will try to explore around the prior. Many methods of exploration have been studied in literature, for example intrinsic motivation \cite{pathakICMl17curiosity}, or maximizing state coverage \cite{fu2017ex2}. Instead, our exploration policy, $\pi_{\text{exp}}$, aims to maximize the change that the agent causes in the environment. Since our actions are close to the prior, it is likely that any changes caused by the agent in the environment will be meaningful and not destructive. Mathematically, for a given video $R_k$ this can be described as: 

\begin{equation}
    \label{eq:expl}
    c_k = \max_{i, j} ||\Phi_f(R_{k, i}) -\Phi_f(R_{k, j})||_2
\end{equation}

where $i$ and $j$ are different frames of the video and $\Phi_f$ is a frame-by-frame embedding of the video. This exploration policy uses the same exact inputs and setup as the task policy, as well as the CVAE architecture. 

As the robot interacts with the environment, we want the task policy to improve to achieve more success. Thus, we need to have a notion of how good the robot is doing compared to the target human video. Given that human and robots have different morphologies, and perform tasks differently, how do we create a representation space for our objective function? 

\subsubsection{Representations for Human-to-Robot Video Alignment}

Trying to learn correspondences between human and robot videos can be a challenging task. Prior works \cite{sharma2019third, Sermanet2017TCN, smith2019avid, sermanet2016unsupervised} have attempted to achieve this via learning paired or unpaired videos from a single scene to learn a joint embedding. This would not scale to large set of in-the-wild manipulation tasks described in Figure~\ref{fig:teaser}. Instead of trying to learn a tight coupling between human and robots, we aim to get a comparison between human and robot video at a \textit{high level}. We postulate that the effect that the agent had on the environment is more important than how the agent moved, since that can vary with different morphologies. We embed both robot and human videos into a space that is agnostic to the agent. We find that inpainting both human and robot from the video allows us to do so. We employ Copy-Paste Networks \cite{lee2019copy}, which, given a mask of an agent, trains a network to copy information from other video frames and inpaint the area masked out. An example of this procedure can be seen in Figure~\ref{fig:inpainting}. 

Only inpainting the human out of the video is not enough to compare the two semantically, since the videos might be of different lengths, different speeds or may have other minor differences. Advances in action recognition \cite{zhang2013actemes, goyal2017something, hara2018can, carreira2017quo, wang2018non, feichtenhofer2019slowfast} have allowed models to determine if two videos which may look different are performing the same task. We use such an action recognition model from \citet{monfort2021multi}, trained on large-scale passive data. This model takes in an entire video, and outputs an embedding. We call the composition of the action recognition model and the inpainting model our \textit{agent-agnostic representation}. We denote this with $\Phi$. Using this agent-agnostic representation, human video $V$ and robot video $R$, our objective function is the distance: 

\begin{equation}
    \label{eq:tasl}
    |\Phi(V) - \Phi(R_{k})||_2
\end{equation}
We use this function to train the task policy. The exploration policy objective (Equation ~\ref{eq:expl}) which is to maximize "change" leverages frame-wise version of $\Phi$, denoted by $\Phi_f$. In order to increase robustness, we sample costs over multiple embeddings with different video-level augmentations. Given a good way to align robot and human videos, how do we optimize our policies in a sample efficient way?

\subsubsection{Sampling-based Optimization Procedure}

RL methods have shown promise in learning by interaction. However, they remain too sample inefficient for large scale robot learning in unstructured settings. Instead, we propose a simple zeroth order sampling based alternative, in a similar fashion to CEM \cite{rubinstein2013cross}. In our sampling procedure, we initially extract the prior from a third person human video $V_k$, and samples residuals:

$$\Delta \Psi_k \sim \mathcal{N}(0, \sigma^2)$$

We execute these $M$ samples $\Psi_{k,m} + \Delta \Psi_{k,m}$ in the real world, and capture resulting videos $R_{k, m}$. We aim to fit $\pi$ to the best performing samples. We repeat this process till convergence. In the following iterations, instead of sampling only from $\mathcal{N}(0, \sigma^2)$, we sample from $\pi$ as well. Using the objective (agent-agnostic) functions described above, we rank trajectories based on these costs and fit the rthjfkcvbjtiuubiuvivkecivnuerujipolicy to the 10 highest ranking $\Delta \Psi_{k, m}$. This set, $E$, is the set of "elites" in CEM. The result of this procedure are trained exploration and task policies. We provide an overview in Algorithm 1.

\section{Experimental Setup}
\label{sec:experiments}

\subsection{Experimental Details}

\vspace{2mm}\noindent\textbf{Hardware$\quad$}
In order to perform manipulation in the wild, a mobile robot is needed. Thus, we use the Stretch Robot \cite{kemp2021design}. This is a mobile base with a 6 dof arm and gripper. Pictures of the hardware setup can be seen in Figure~\ref{fig:setup}. We use a Cartesian position control to command the translation of the wrist, and an in-built orientation controller for the rotation. The default gripper comes with suction cups as fingertips. Our setup uses an Intel Realsense D415 depth camera.

\begin{figure*}[t]
\centering

\begin{subfigure}[b]{0.325\linewidth}
    \includegraphics[width=\linewidth]{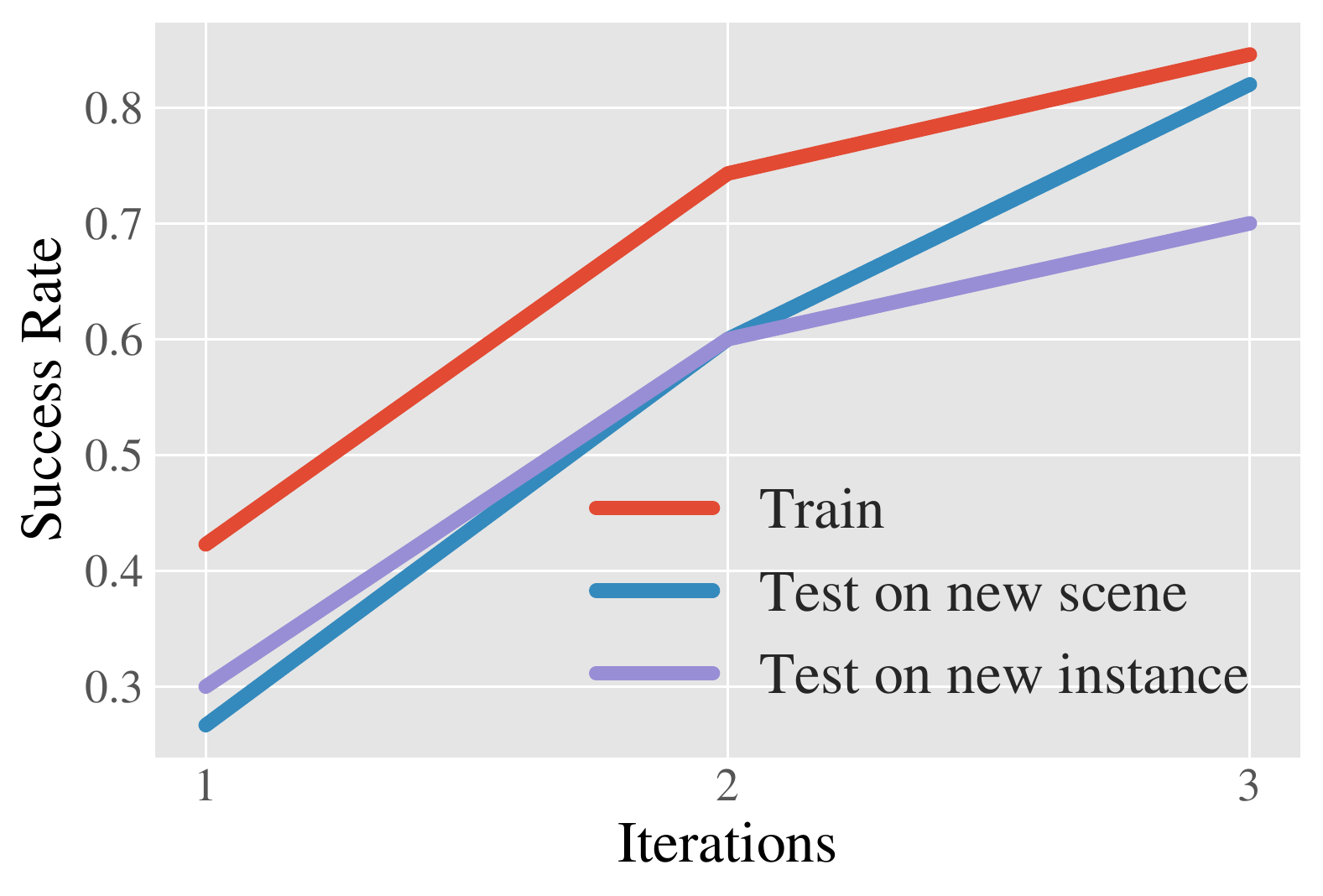}
    \vspace{-0.14in}
    \caption{\small Drawer}
    \label{fig:res-dw}
\end{subfigure}
\begin{subfigure}[b]{0.325\linewidth}
    \includegraphics[width=\linewidth]{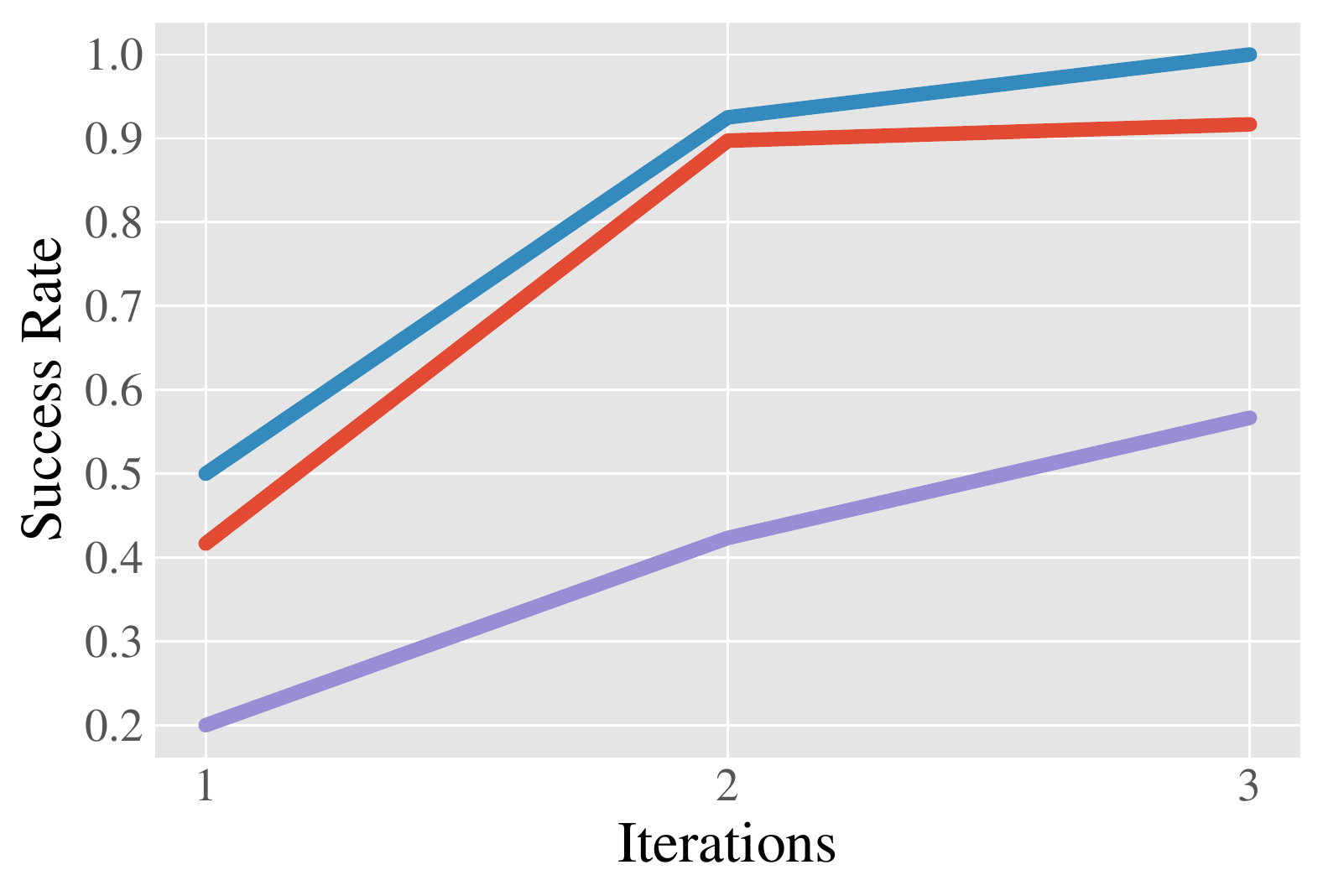}
    \vspace{-0.14in}
    \caption{\small Door}
    \label{fig:res-dr}
\end{subfigure}
\begin{subfigure}[b]{0.325\linewidth}
    \includegraphics[width=\linewidth]{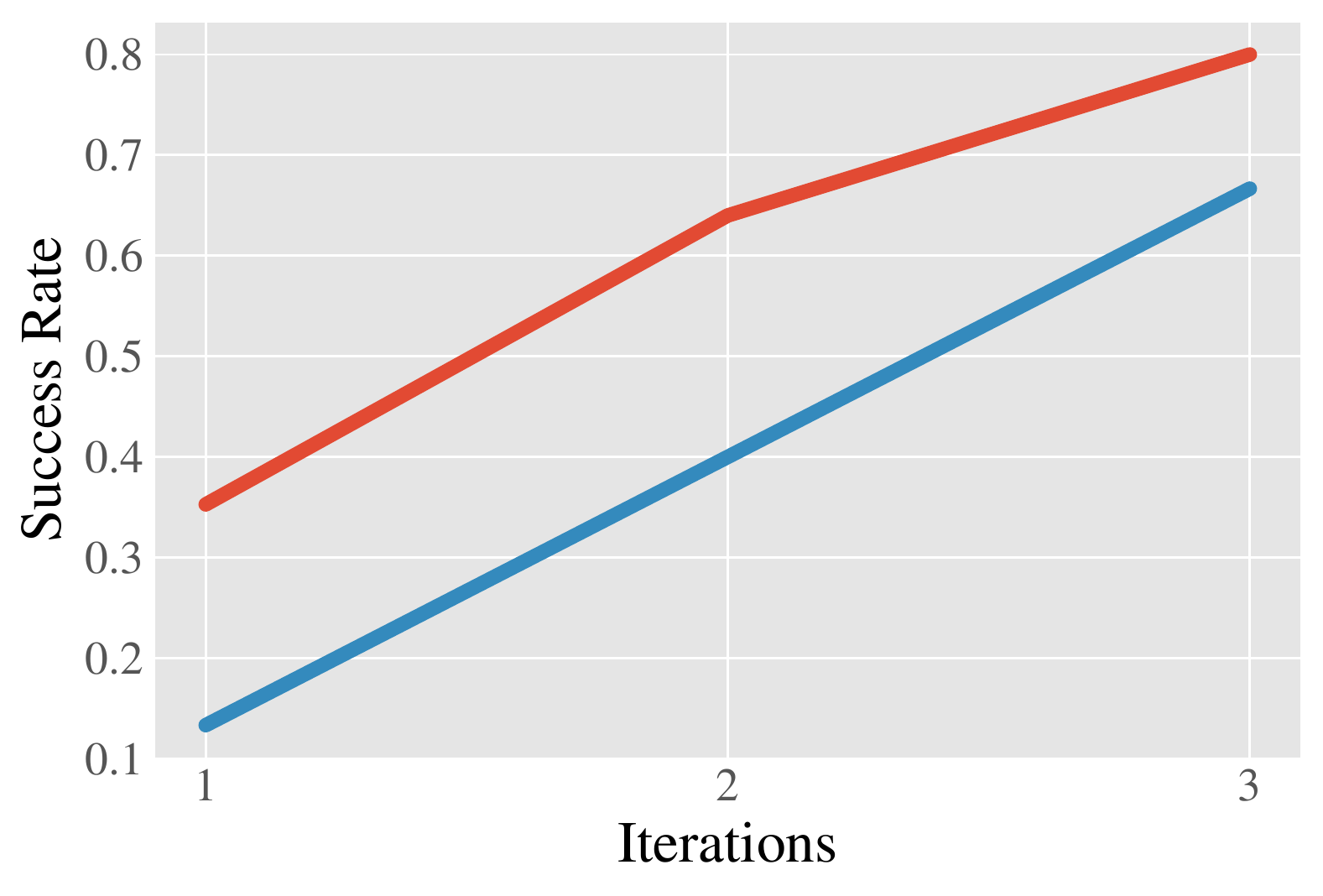}
    \vspace{-0.14in}
    \caption{\small Dishwasher}
    \label{fig:res-ds}
\end{subfigure}
\begin{subfigure}[b]{0.325\linewidth}
    \includegraphics[width=\linewidth]{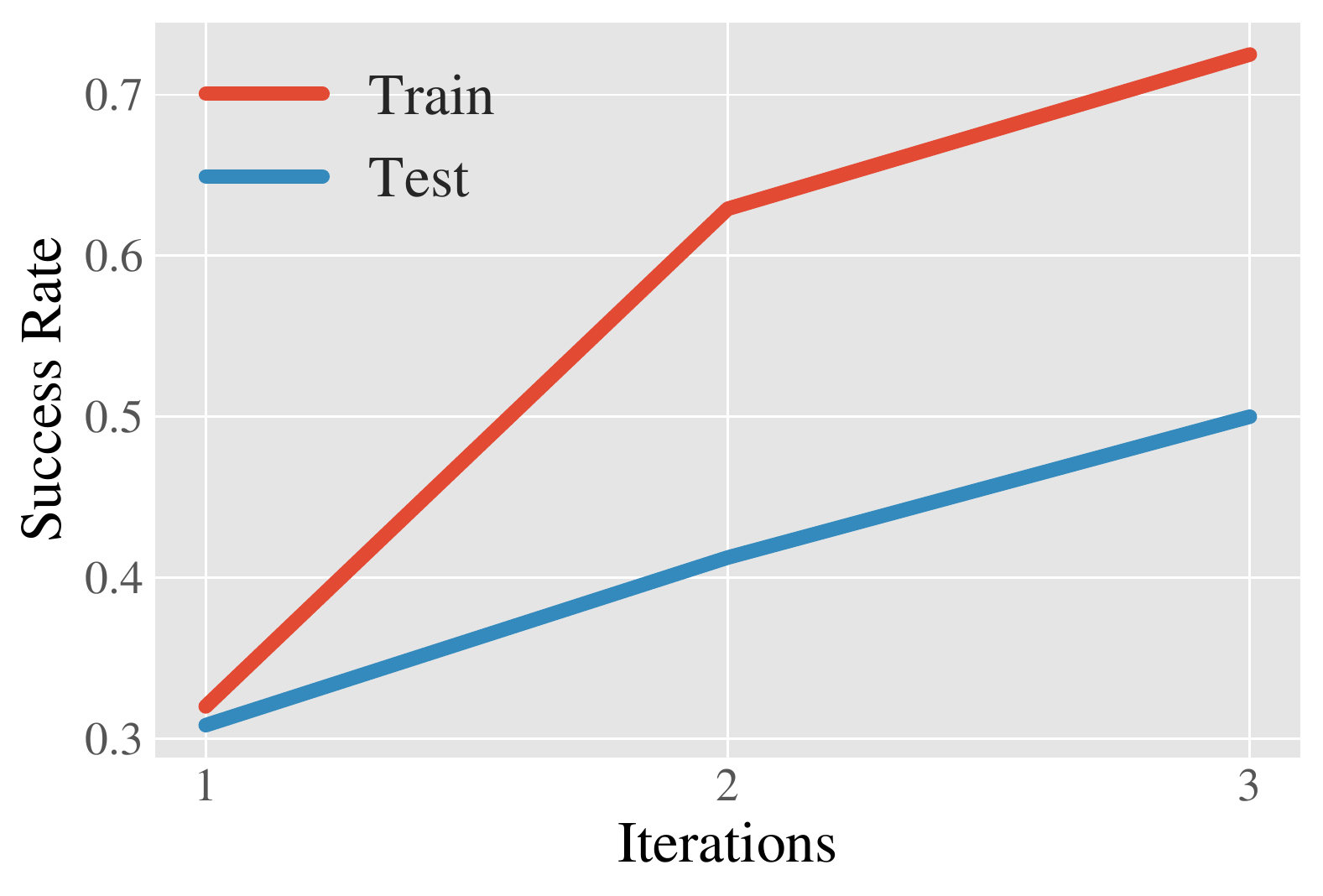}
    \vspace{-0.14in}
    \caption{\small Object in Shelf}
    \label{fig:res-obj}
\end{subfigure}
\begin{subfigure}[b]{0.325\linewidth}
    \includegraphics[width=\linewidth]{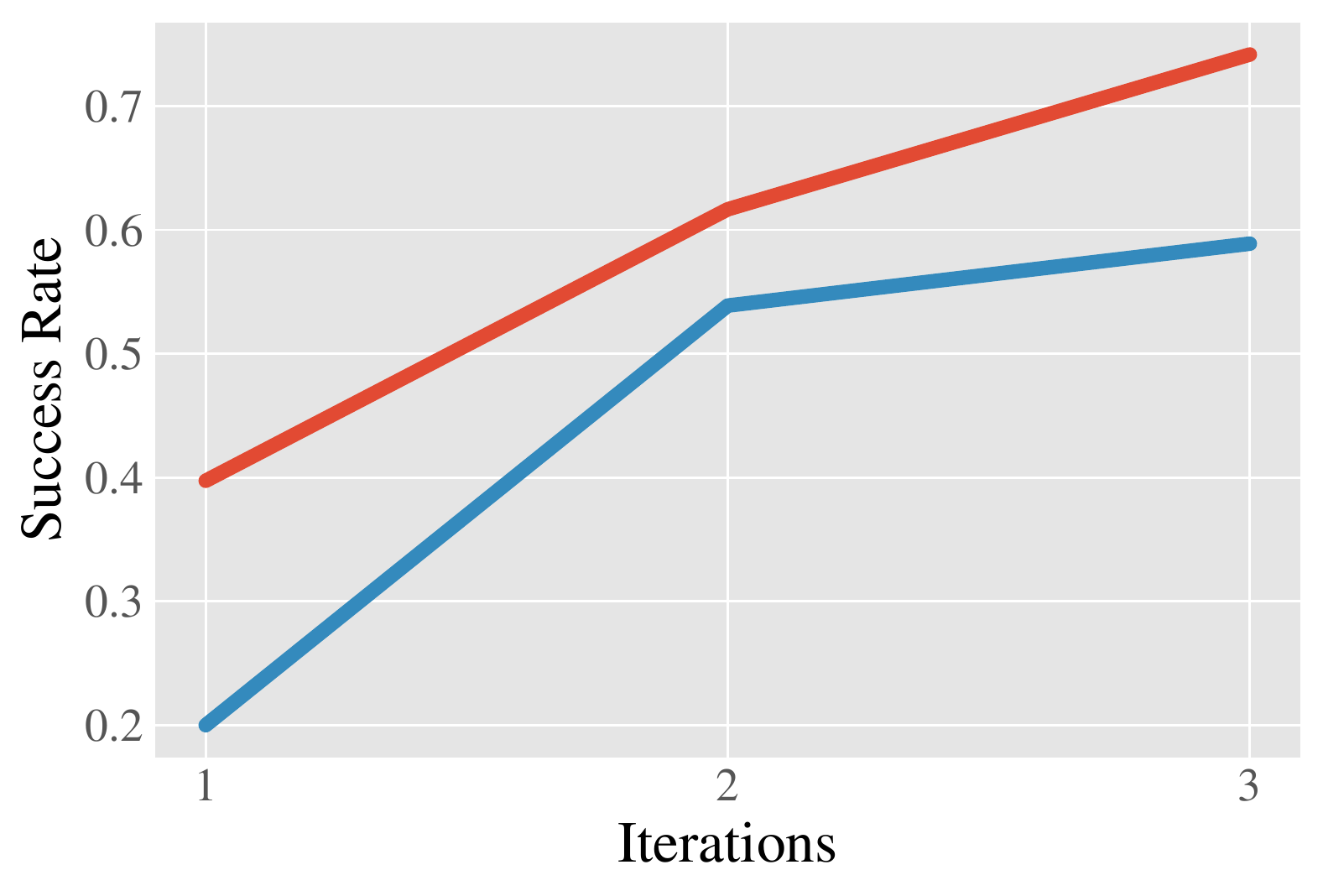}
    \vspace{-0.14in}
    \caption{\small Multi-task}
    \label{fig:res-multi}
\end{subfigure}
\begin{subfigure}[b]{0.325\linewidth}
    \includegraphics[width=\linewidth]{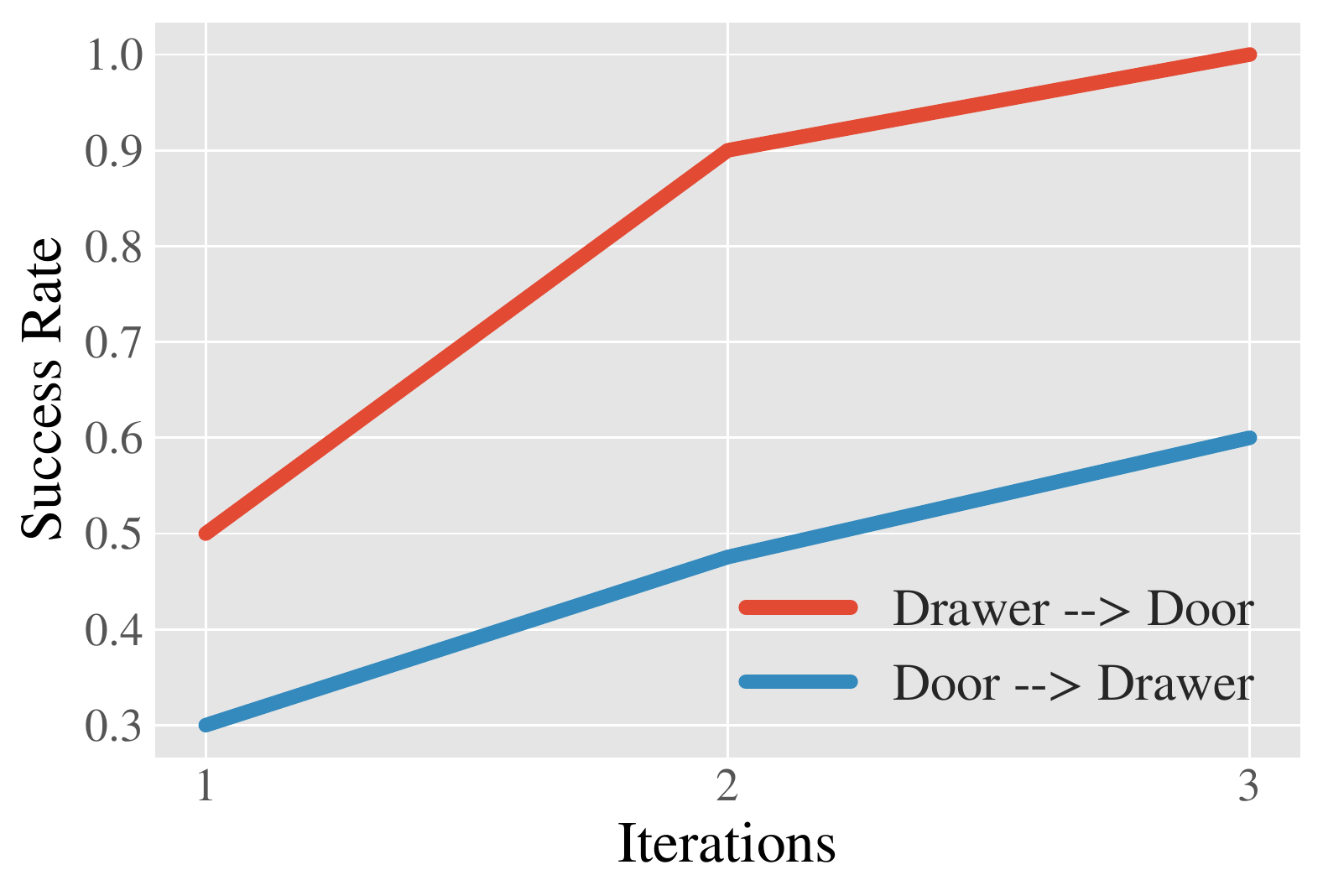}
    \vspace{-0.14in}
    \caption{\small Task-level Generalization}
    \label{fig:res-gen}
\end{subfigure}

\vspace{-0.04in}
\caption{\small We present results of our thorough investigation of \ours on various kitchen tasks such as drawer (a), door (b) and dishwasher (c) opening and closing, a task involving picking and placing different objects into shelves (d). We test multi-task policies (e) trained on a subset of the tasks and generalization between tasks (f). We report training and testing success rates (out of 1) for 3 iterations of training. }
\vspace{-0.12in}
\label{fig:res}
\end{figure*}

\vspace{2mm}\noindent\textbf{ Environment and Data Collection$\quad$}
We perform experiments on every-day objects and settings, for example drawers, dishwashers, fridges in different kitchens, doors to various cabinets. Data collection is performed in various in the wild settings. Each of the tasks presented in Figure~\ref{fig:res} was trained over three demonstrations.  Our setup involves 20 tasks seen in Figure~\ref{fig:teaser} and in the Appendix.

\subsection{Baselines and Ablations}

Our method relies on several key ideas, such as policy learning from interactions, the high-level video alignment to compare robot trajectories and human demonstrations, and the task-agnostic policy we use. We test both the performance and sensitivity of \ours to design choices. We compare against several state-of-the-art baselines. Performing RL in the real world is infeasible \cite{zhu2020ingredients} for our large set of tasks and in diverse settings. Thus we compare against offline RL, which learn interaction policies from data, and do not interact with the environment at training time. We modify a SOTA method in offline RL, Conservative Q-learning (CQL) \cite{kumar2020conservative} to work with our extracted human priors. The policy predicts the residual to the prior, just like ours, even using the same inputs as well. The reward function used is the negative L2 error between the 3D ResNet \cite{feichtenhofer2019slowfast} embedding of target and robot videos. This baseline is trained with same number of samples as \ours (30 samples x 2 training iterations). We call this approach \texttt{CQL}. We also train \texttt{CQL} with the same objective function as \ours (agent-agnostic). We call this baseline \texttt{CQL-ours}. We compare against competing SOTA approaches for learning joint human and robot embedding spaces. We train a Time Contrastive Network (TCN) \cite{Sermanet2017TCN} to extract a representation from human and robot videos. We call this baseline \texttt{CQL-TCN}. The reward used for CQL is the the distance (between human and robot videos) in TCN embedding space. Similarly, Cycle-GAN \cite{zhu2017unpaired} has been shown to be useful translating robot and human demonstrations \cite{smith2019avid}. In the same fashion as \citet{smith2019avid}, we employ Cycle-GAN representations (trained on both human and robot videos) as an embedding space for the reward function. We refer to this baseline as \texttt{CQL-CycleGAN}. Finally, we compare to an off-the-shelf implementation of Behavior Cloning (\texttt{BC}). This approach is similar to \ours, however is without iterative refinement. A key difference is that the policy is trained with a standard L2 loss on the output actions, unlike our VAE-based policy.

\section{Results}
\label{sec:results}

We evaluate \ours in various real world, in-the-wild settings in order to answer the following questions: 

\begin{itemize}
    \item Can \ours work for a large set of in-the-wild robot manipulation tasks? 
    \item How can we use \ours to generalize to new scenes, objects and settings? 
    \item How much do the individual components (i.e. policy learning and agent agnostic cost function) of \ours help?
    \item How does \ours compare agents SOTA approaches? 
\end{itemize}

We attempt to answer these via various experiments on in-the-wild tasks, analyzing generalization capabilities of \ours on a few tasks, comparing to competing methods, and addressing the question of the importance of the individual components of our method, such as iterative improvement, our agent-agnostic objective and the exploration policy. In Figure~\ref{fig:res} we present the results of our experiments on four tasks: opening and closing a drawer, door and dishwasher, as well as placing different objects in shelves. The setup is as described in Section~\ref{sec:experiments}.

\subsection{Robot Learning in the Wild}

We provide results on a large set of 20 tasks, ranging from turning on a water tap to folding a shirt. Images of these tasks can be seen in Figure~\ref{fig:teaser} or in the Appendix. We show that \ours is able to to scale to a wide variety of tasks that involve large fixed objects such as fridges, or smaller rigid objects such as our ball-in-hoop task, or handling soft objects such as our shirt folding and whiteboard cleaning tasks. We are able to train these in only a few hours, in diverse locations and settings. We provide videos of these tasks in the supplementary material and at \texttt{\url{https://human2robot.github.io}}.

\subsection{Evaluation and Comparison to Baselines}

\vspace{2mm}\noindent\textbf{ Tasks$\quad$}
We perform a drawer opening and closing task in a kitchen. We report results averaged across three human demonstrations on two drawers (Figure~\ref{fig:env-dw}), and 3 iterations of \ours. Similarly to the drawer task, we perform a door opening and closing task. Figure~\ref{fig:env-dr} shows the task setup. We train on two doors and test on the third one. We perform 3 iterations of \ours. The third task involves opening and closing a dishwasher, as shown in  Figure~\ref{fig:env-ds}. Due to a lack of dishwashers in the kitchens, we only train on one dishwasher, and test on a held-out dishwasher in a different scene. For the fourth task we train a policy for picking and placing four objects in three different shelves. These objects vary in size and shape (bottles, cans, etc.). We provide a single demonstration for each object. We test on two held-out objects and a held-out shelf placement (starting point is top shelf and goal is middle shelf). Details about the task can be found in the Appendix.

\vspace{2mm}\noindent\textbf{Training Evaluation$\quad$}
We look into the (training) results running \ours on the various tasks described above. In Figure~\ref{fig:res-dw}, we show a learning curves of success rates over 3 iterations of \ours for the drawer task. The training curve (red) shows an increase from about a 43\% success rate to 83\% success rate after two iterations. The initial success rate is the success rate of the prior. We see a clear improvement during training with \ours. For the door task (Figure~\ref{fig:res-dr}, red), the training curve shows an increase in performance from about 40\% success to 92 \% success, indicating that \ours is able to learn to improve iteratively. The training curve for the dishwasher task (Figure~\ref{fig:res-ds}, red) shows that \ours also improves for this task, similarly to the drawer and door tasks. Figure~\ref{fig:res-obj} provides learning curves for the task of picking and placing objects from shelves. We can see that the the train (red) curves show improvement. However, we did note that this task required a lot more precision than the kitchen tasks presented above, which is why we see the test success rate to be much lower than the train one. We often found that if the robot predicted the waypoints incorrectly by even a couple of centimetres, it would hit a shelf and get stuck.

\vspace{2mm}\noindent\textbf{ Comparison to Baselines$\quad$}
We compare \ours to both offline RL and Behavior Cloning. We present results of multiple instances of offline RL in Table~\ref{tab:res-table} (we report and compare training results). All methods had the same amount of data to train on. We report success rates out of 1. Our method strongly outperforms all the baselines. Offline RL, especially with smaller datasets, has difficulty in learning without any online interaction. Learning both actor and critic require data that covers more of the action space than is likely available in the wild. Interestingly, \texttt{CQL-ours} tends to outperform other approaches such as \texttt{CQL-CycleGan} (similar to that presented by \citet{smith2019avid})  \texttt{CQL-TCN} (similar to \citet{Sermanet2017TCN}), and \texttt{CQL} \cite{kumar2020conservative}. Similarly, Behavior Cloning, which uses our agent-agnostic cost function to filter the top trajectories, mostly outperforms the offline RL approaches. This indicates that our objective function is able to differentiate between good and bad trajectories in many algorithmic settings.  

\begin{table}[t]
\centering
\resizebox{0.85\linewidth}{!}{%
\begin{tabular}{lcc}
\toprule
 & Drawer & Door \\
\midrule
\multicolumn{2}{l}{\textit{No iterative improvement}:}\vspace{0.4em}\\
Behavior Cloning & 0.53 & 0.30  \\
Offline RL (\texttt{CQL-ours}) & 0.47 & 0.30 \\
Offline RL (\texttt{CQL-CycleGAN}) \cite{smith2019avid} &  0.23 & 0.30 \\
Offline RL (\texttt{CQL-TCN}) \cite{Sermanet2017TCN} &  0.27 & 0.20 \\
Offline RL (\texttt{CQL}) \cite{kumar2020conservative} & 0.33 & 0.13 \\
\midrule
\multicolumn{2}{l}{\textit{No agent-agnotic objective}:}\vspace{0.4em}\\
\ours (ours) &  0.47 &  0.53   \\
\midrule
\multicolumn{2}{l}{\textit{No Exploration Policy}:}\vspace{0.4em}\\
\ours (ours) &   0.60 &  0.73  \\
\midrule
\textbf{\ours (ours)}  & \textbf{0.83} & \textbf{0.92} \\
\bottomrule
\end{tabular}}
\caption{\small We present a set of evaluations on two real world tasks: drawer and door. The task is to imitate the third person demonstration of the human. The results presented our averaged over 30 trials.}
\label{tab:res-table}
\vspace{-0.15in}
\end{table}

\begin{figure*}[t]
\centering

\begin{subfigure}[b]{0.325\linewidth}
    \includegraphics[width=\linewidth]{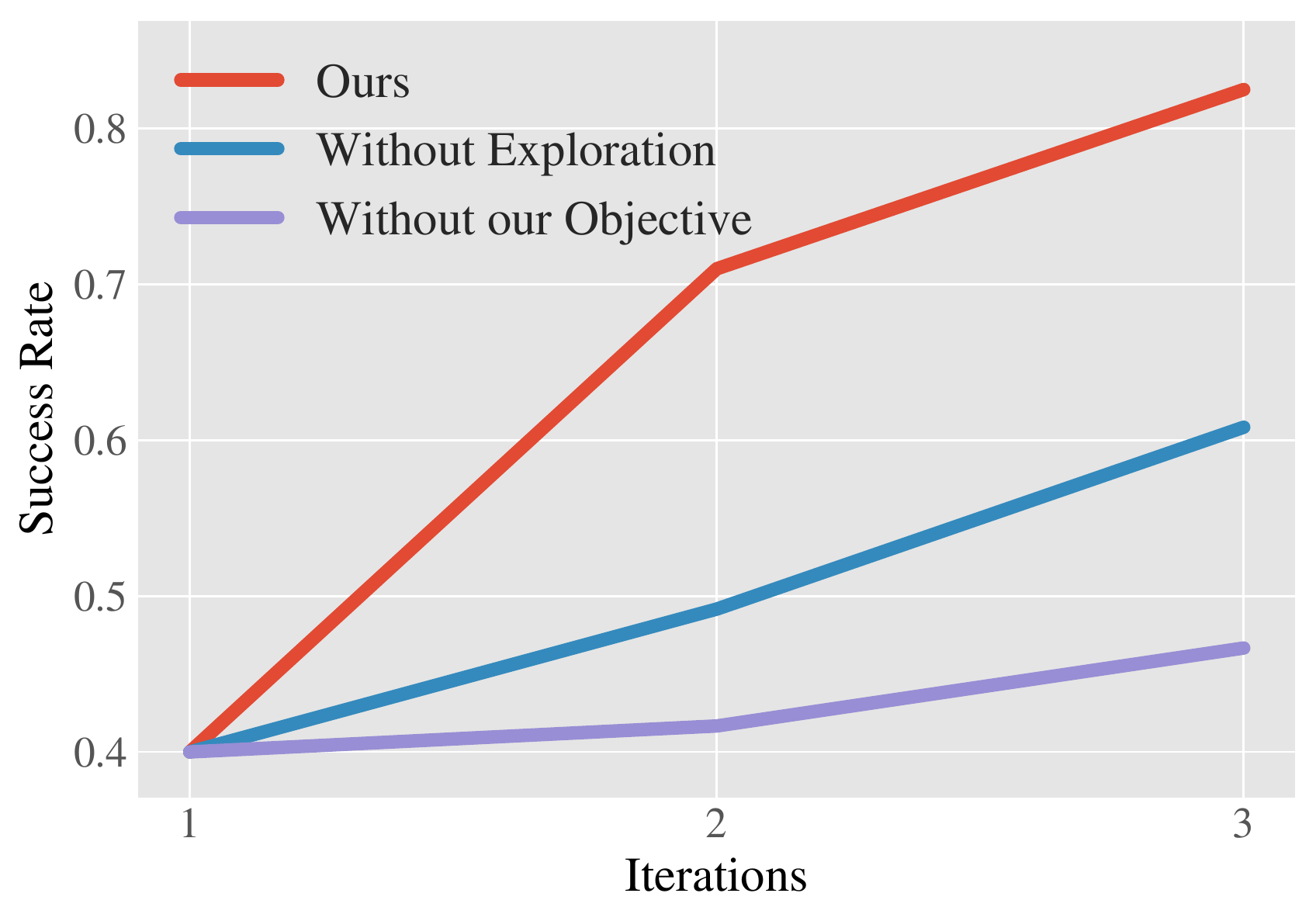}
    \vspace{-0.13in}
    \caption{\small Drawer}
    \label{fig:abl-dw}
\end{subfigure}
\begin{subfigure}[b]{0.325\linewidth}
    \includegraphics[width=\linewidth]{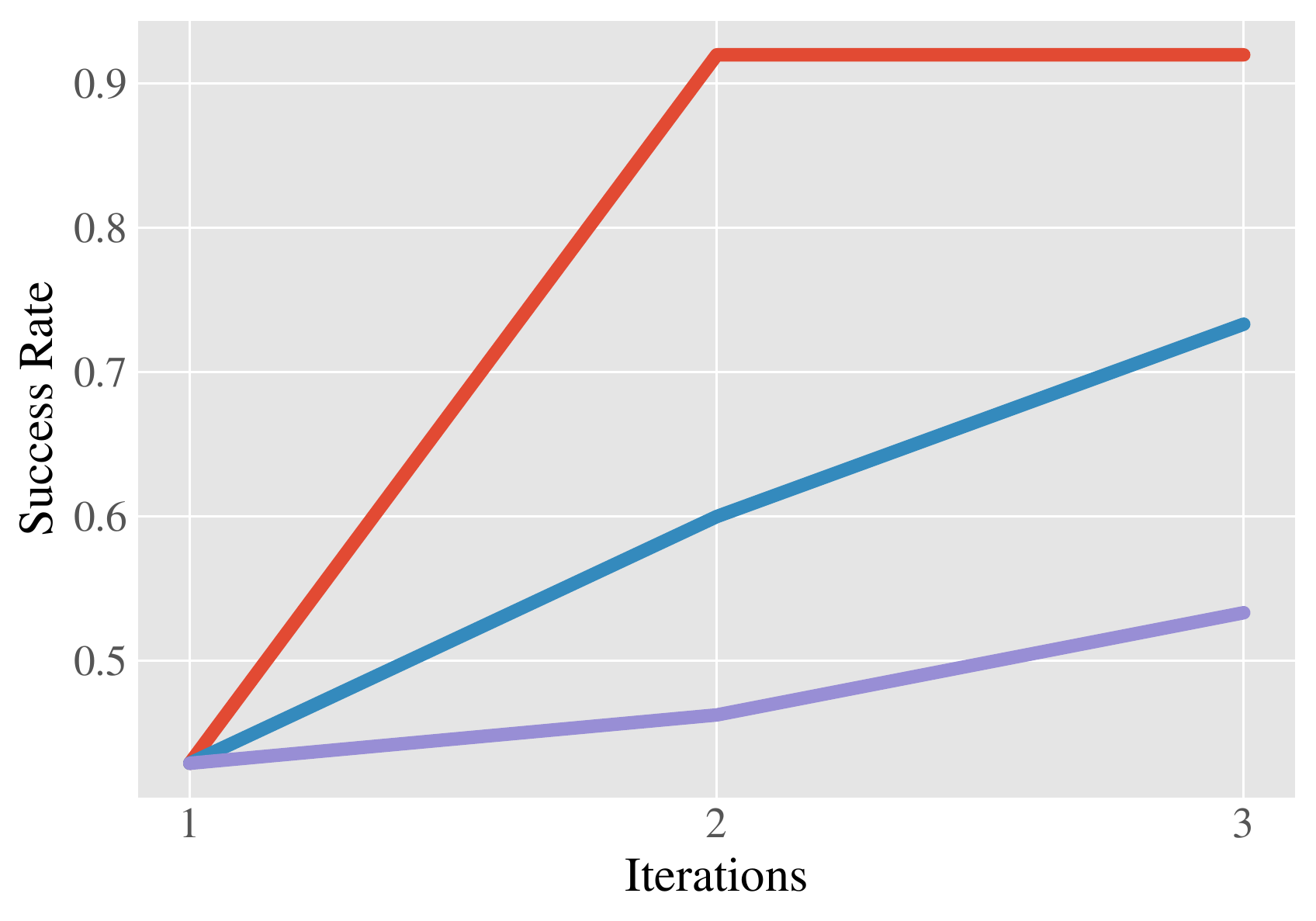}
    \vspace{-0.13in}
    \caption{\small Door}
    \label{fig:abl-dr}
\end{subfigure}
\begin{subfigure}[b]{0.325\linewidth}
    \includegraphics[width=\linewidth]{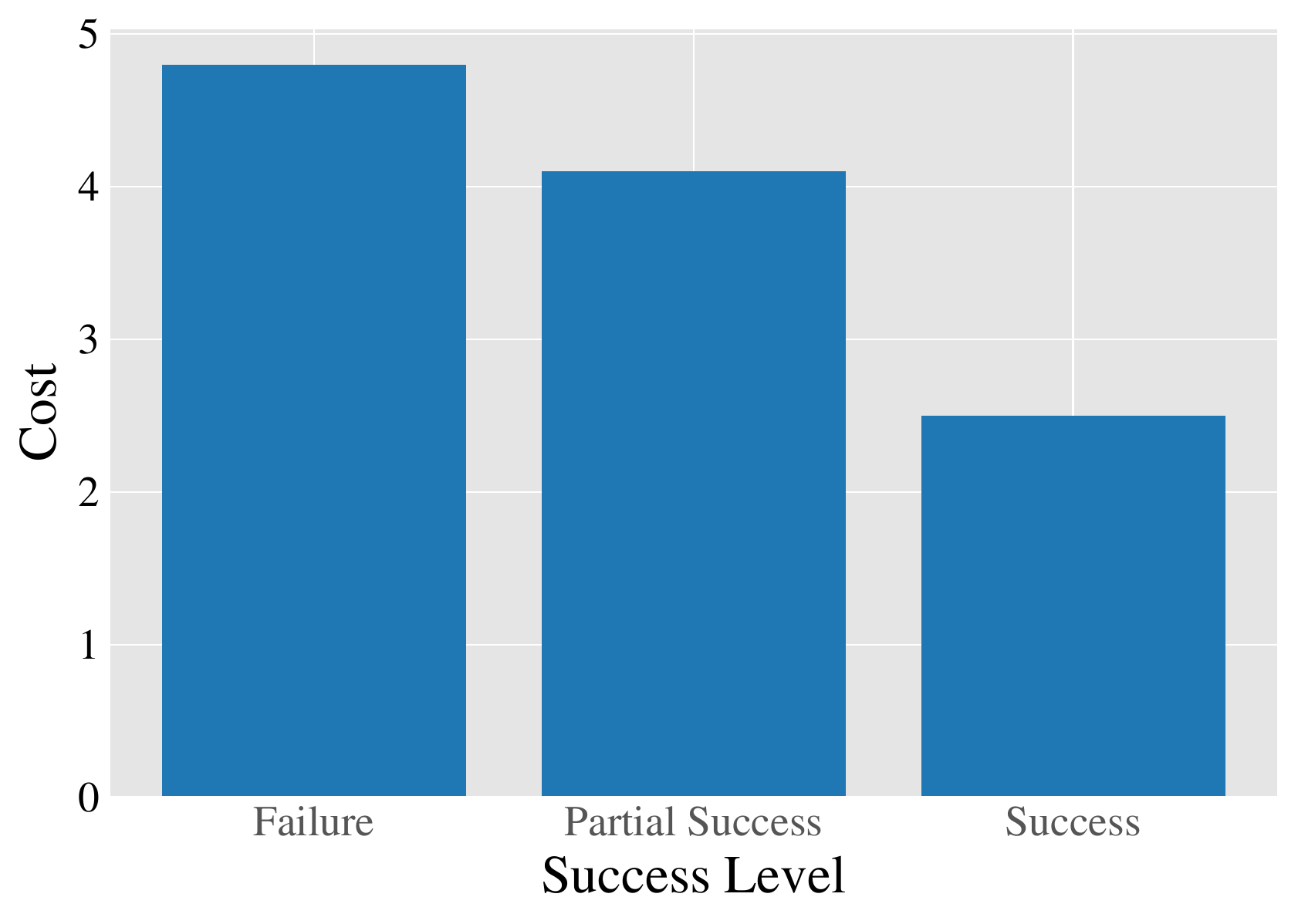}
    \vspace{-0.13in}
    \caption{\small Cost Function}
    \label{fig:abl-rew}
\end{subfigure}
\vspace{-0.04in}
\caption{\small We ablate different aspects of \ours. We analyze the need for our task-agnostic exploration policy, as well as our agent-agnostic representations. (a) is the drawer task, (b) is the door task. In (c) we analyze on the drawer task how our cost functions compare for different levels of success obtained by a trajectory.}
\vspace{-0.17in}
\label{fig:abl}
\end{figure*}

\subsection{Generalization to New Instances}

We also evaluate how well policies trained with \ours are able to perform on new instances of the same task that they were trained on. In the drawer task, we test the policy on a held-out drawer. The learning curve for the held-out drawer (Figure~\ref{fig:res-dw}, blue) shows that the achieved success is lower than that on the training drawers. The detected prior for each demonstration may have different biases and errors. A policy may not necessarily transfer as well to a new demonstration without any training, however, as we see here, we expect improvement as there are commonalities in the structure of the task. Another common aspect is the camera and geometry information for both train and test demonstrations (since the view is the same). For the door task (Figure~\ref{fig:res-dr}, blue), interestingly, the success rate is higher here than the train one. As discussed previously, we expect that this is due to the strength of the prior for this specific door. For the object task, The policy was tested on two held objects and held out shelf placements. The test (blue) curves in Figure~\ref{fig:res-obj} shows a definite improvement in success rate over multiple iterations. From these experiments, we see that \ours does have the ability to generalize to new instances of the same task it was trained on.

\subsection{Generalization to New Scenes:} 

To analyze how well \ours performs in a new scene, where the calibration and geometry are different, we try the trained policy on a drawer in a different part of the kitchen. Note that the camera angle and view also change. The resulting curve is shown in Figure~\ref{fig:res-dw} (purple). We see that similar to the test curve on the held-out drawer, there is definitely an increase in the success rate, however, the performance is still worse than the train drawers. In the case of the door task (Figure~\ref{fig:res-dr}, purple) the performance of this policy in a new setting is significantly worse, unlike the drawer task,. This is likely due to mismatches in the train and test demonstrations. Although the performance is not as strong, there is still an improvement in success rate from 20\% to about 57\%. For the dishwasher task (Figure~\ref{fig:res-ds}, blue) we see a strong improvement in the success rate, similarly to the other two tasks. As expected, the policy does not perform as well as it does on the train dishwasher. We see that \ours allows for generalization to new scenes, however, in most cases the performance is worse than running \ours on new instances, most likely due to large visual changes, as well as different geometry and calibrations.

\subsection{Generalization to New Settings:}

In order to test the generalization between tasks, we test the trained policy on a held-out door (Figure~\ref{fig:env-dr}), in order to test the task level generalization. We see a strong performance on the door, as shown in Figure~\ref{fig:res-gen} (red). We see generalization from the drawer policy to the door task. We suspect that the final high success rate is due to the fact that the prior for this specific door was much more accurate than for other instances or tasks. Nevertheless, we see an improvement on the held-out door as the policy is trained more, indicating that \ours is able to improve the performance for not only the task at hand, but also allows for some degree of generalization across tasks. For transferring the policy trained on the door task to the drawer task, we see in Figure~\ref{fig:res-gen} (blue) an improvement in performance on the drawer task, similarly to the policy trained on the drawer task. This definitely indicates that there is some degree of task-level generalization in \ours.

\subsection{Multi-task Generalization}

In the previously described experiments, we trained policies for one task only, even if it was tested on another one. With this experiment, we aim to answer how training a joint policy would work. Using a similar approach as described above for each of the 3 tasks (drawer, door and dishwasher) we train the same policy $\pi$. We test this policy on the same test scenes as the previous experiments. We present results in Figure~\ref{fig:res-multi}. We can see clear improvement over all the iterations, both in the training and testing instances. However, the success rates are lower than policies trained for individual tasks which makes sense. We see a big increase from the first iteration. This is likely because there is more data thus generalization becomes a little easier as compared to individual policies. However, as more training happens this does not remain the case.

\subsection{Sensitivity of \ours} 

We test the sensitivity of our method to both the inpainting process and the exploration policy. Firstly, We train \ours without the agent-agnostic representations. Secondly, We also compare \ours against a version which does not use the exploration policy.  In Figure~\ref{fig:abl} we perform ablations to test the sensitivity of \ours to various components. In previous experiments, we have seen that the iterative improvement provides a lot of benefit, as evidence by the increase in performance over iterations in almost all of the tasks and scenarios shown in Figure~\ref{fig:res}, as well as the boost in performance over offline RL methods, as shown in Table~\ref{tab:res-table}.

\vspace{2mm}\noindent\textbf{Agent-Agnostic Objective $\quad$}
We train our policy without our agent-agnostic objective function, on both the drawer and door tasks, as shown in  Figure~\ref{fig:abl-dw} and \ref{fig:abl-dr} (purple). We see almost no gain in success rate from the initial samples from the prior in either task. This is likely due to the fact that video alignment models focus too much on the agents, thus the true top trajectories might not be selected. We conclude that an agent-agnostic objective is crucial to the success of \ours.

\vspace{2mm}\noindent\textbf{ Exploration$\quad$}
We test a version of \ours without our exploration policy. We see this in Figure~\ref{fig:abl-dw} and\ref{fig:abl-dr}, in the blue curve. While the performance of this version of \ours outperforms the ablation that does not use agent-agnostic objective, we still see a drop in success rate from the \ours with the exploration policy. This shows that the method can learn without biasing exploration around maximizing "change" in the environment, but it will be slower. 

\vspace{2mm}\noindent\textbf{Analyzing our Objective Function $\quad$}
In Figure~\ref{fig:abl-rew}, we show a bar plot of distances in our agent-agnostic embedding space for the drawer task. We present the cost for three types of trajectory. Note that each category is averaged over multiple trajectories. "Failure" trajectories completely fail, and the robot never touches the drawer. "Partial Success" trajectories open the wrong drawer or do not open/close the drawer fully. "Success" trajectories match the human demonstrations. We see that there the cost decreases between these three and that there is big drop between "Partial Success" and "Success". Since the cost is not close to 0, we expect there to be noise in the measurement of video alignment. However, empirically we found that our embedding space was robust enough to differentiate between successful and unsuccessful trials. In future work, we hope to train such an embedding space.

\section{Conclusion} 
\label{sec:discussion}
We propose \ours, an efficient real-world robot learning algorithm that can learn manipulation policies in-the-wild from human videos. Our method leverages advances in computer vision to understand human videos and obtain priors such as hand interactions, movement and direction. \ours is able to efficiently improve in the real world by using our sampling-based policy optimization strategy and agent-agnostic representations, as well as our proposed exploration strategy that maximizes the changes seen in the environment. We perform a thorough evaluation in terms of absolute performance, comparison to SOTA baselines, and generalization to new tasks and scenes on multiple tasks on real kitchens and see strong results. We show that our method is able to work on 20 different tasks in the wild (outside lab settings). \ours is a first step towards learning robot skills from watching internet videos. In the future, we hope to build upon our method to try to solve tasks without online interactions and human demonstrations, i.e. directly learn policies from passive offline video datasets such as YouTube.

\section*{Acknowledgments}
We thank Jason Zhang, Yufei Ye, Aravind Sivakumar, Sudeep Dasari and Russell Mendonca for very fruitful discussions and are grateful to Ananye Agarwal, Alex Li, Murtaza Dalal and Homanga Bharadhwaj for comments on early drafts of this paper. AG was supported by ONR YIP. The work was supported by Samsung GRO Research Award, NSF IIS-2024594 and ONR N00014-22-1-2096. 

\bibliographystyle{plainnat}
\bibliography{main}

\newpage 
\appendix
\subsection{Videos}

Videos of our results can be found at: 
\begin{itemize}
    \item \texttt{\url{https://human2robot.github.io}}
    % \item \url{https://drive.google.com/file/d/1c1pu881OgejetqRrrBDrjY7RID8ZbI8n/view?usp=sharing}
\end{itemize}

\subsection{Implementation Details}

\subsubsection{Real Robot Setup}

We use use the Stretch robot from \citet{kemp2021design}. This is a robot with a mobile base and a wrist with suction cups as fingertips. All 6 degrees of the freedom of the robot are controllable, and we use code provided in \url{https://github.com/hello-robot} to control the robot. To capture videos of humans and the robot we use an Intel Realsense D415. We obtain both depth and RGB images from this camera setup. Each human demonstration takes about 30 seconds. Similarly, each robot episode also takes 30-45 seconds. Overall training takes anytime between 4 and 6 hours, including time to compute inpainted videos (which is the bottleneck in terms of software). 

\begin{figure}[h!]
\centering
\begin{subfigure}[b]{0.47\linewidth}
    \includegraphics[width=\linewidth]{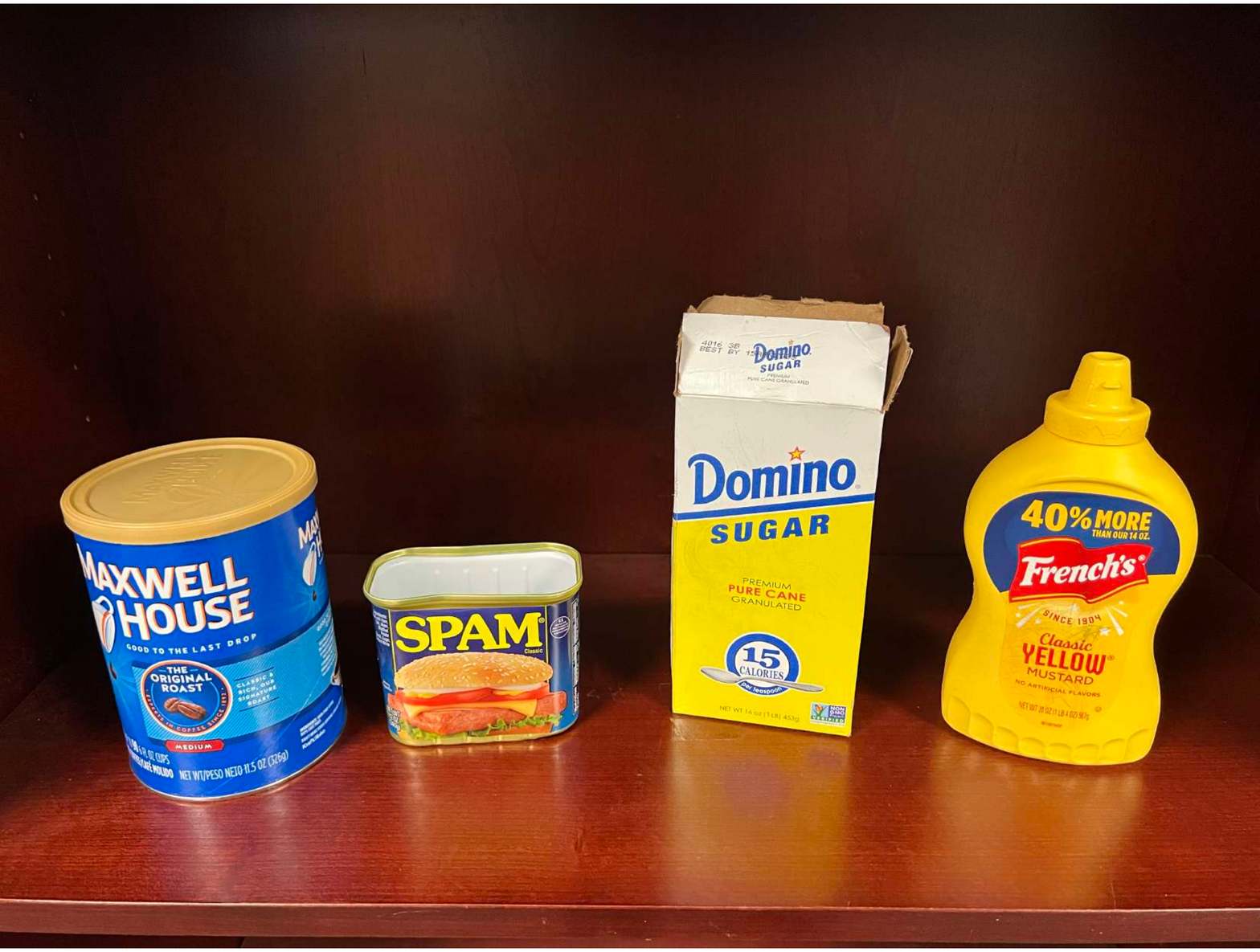}
    \caption{\small Training Objects}
    \label{fig:train-obj}
\end{subfigure}
\begin{subfigure}[b]{0.47\linewidth}
    \includegraphics[width=\linewidth]{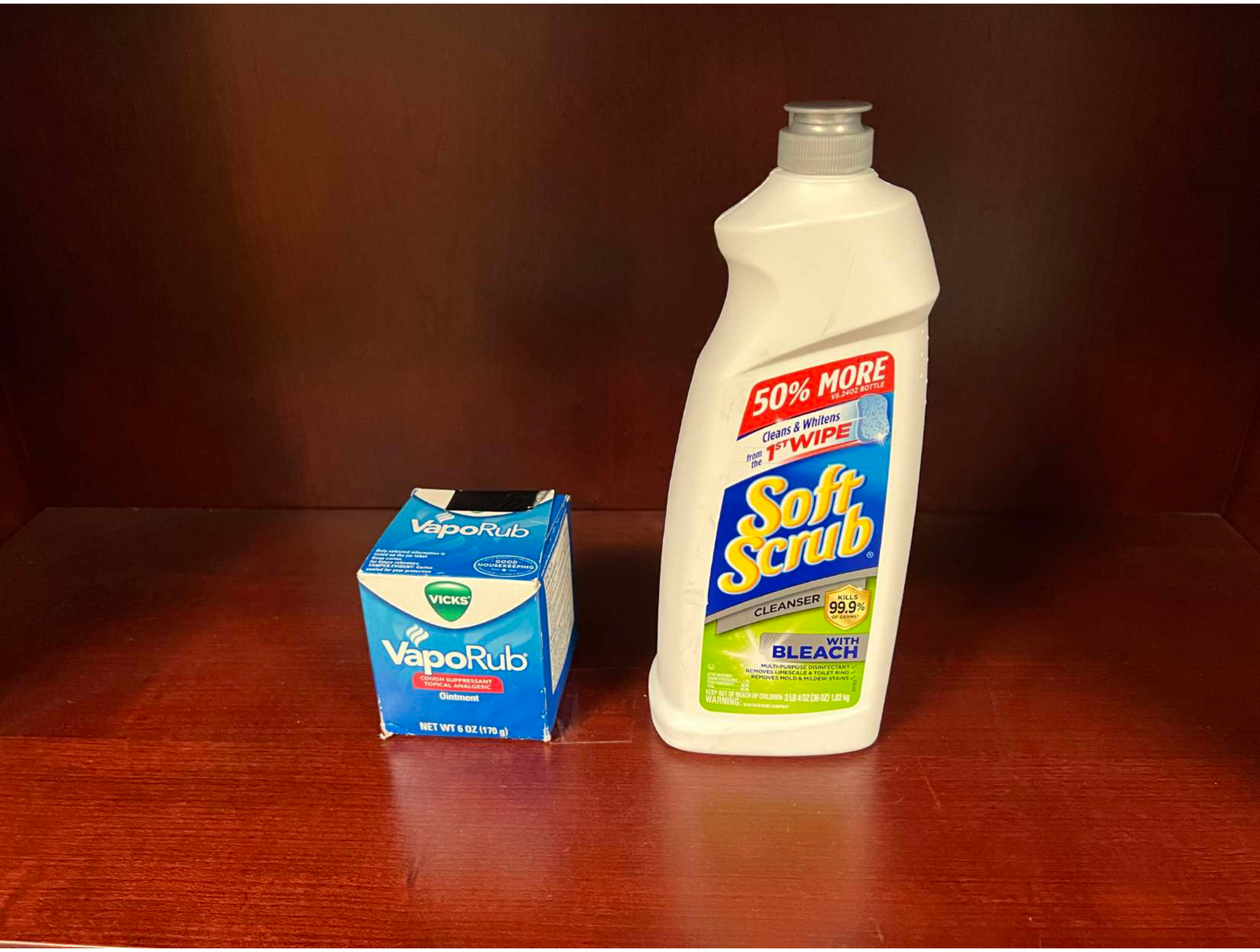}
    \caption{\small Test Objects}
    \label{fig:test-obj}
\end{subfigure}
\caption{\small Images of the objects we used in our shelf pick-and-place task. Objects a-d are train objects, objects e and f are test objects}
\vspace{-0.1in}
\label{fig:objects-split}
\end{figure}

\begin{figure*}[ht!]
\centering
\begin{subfigure}[b]{0.24\linewidth}
    \includegraphics[width=\linewidth]{figs/env_dw.png}
    \caption{\small Drawer}
\end{subfigure}
\begin{subfigure}[b]{0.24\linewidth}
    \includegraphics[width=\linewidth]{figs/env_dr.png}
    \caption{\small Door}
\end{subfigure}
\begin{subfigure}[b]{0.24\linewidth}
    \includegraphics[width=\linewidth]{figs/env_dish.png}
    \caption{\small Dishwasher}
\end{subfigure}
\begin{subfigure}[b]{0.24\linewidth}
    \includegraphics[width=\linewidth]{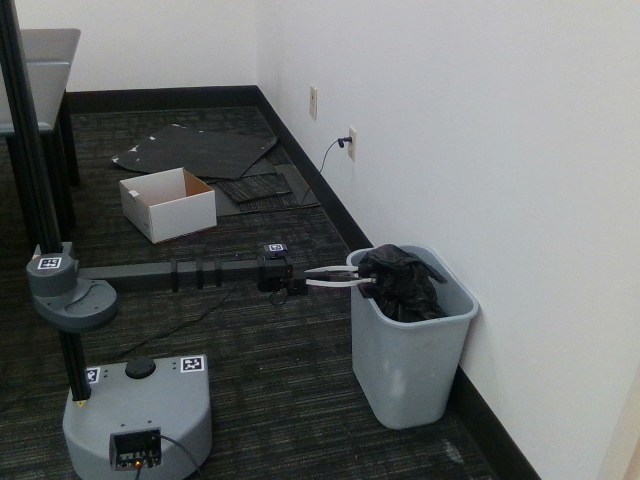}
    \caption{\small Pulling Garbage Bag}
\end{subfigure}
\begin{subfigure}[b]{0.24\linewidth}
    \includegraphics[width=\linewidth]{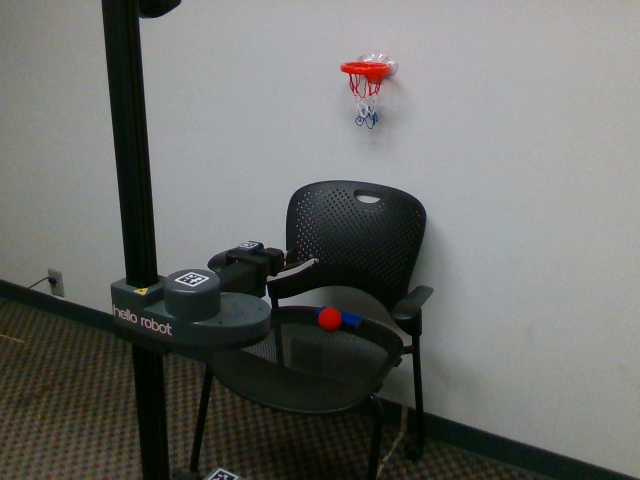}
    \caption{\small Ball in Hoop}
\end{subfigure}
\begin{subfigure}[b]{0.24\linewidth}
    \includegraphics[width=\linewidth]{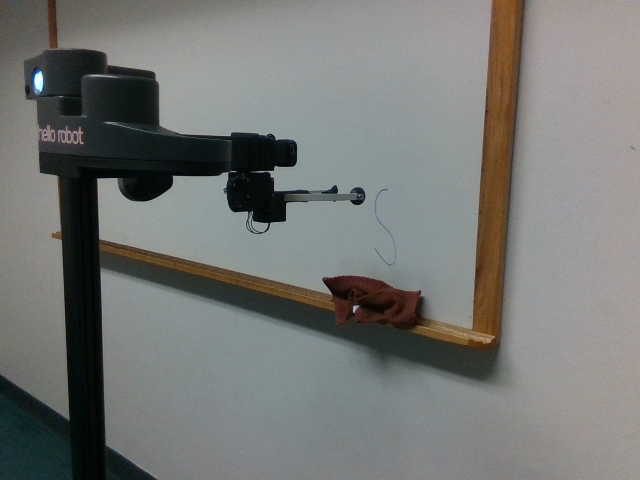}
    \caption{\small Cleaning Whiteboard}
\end{subfigure}
\begin{subfigure}[b]{0.24\linewidth}
    \includegraphics[width=\linewidth]{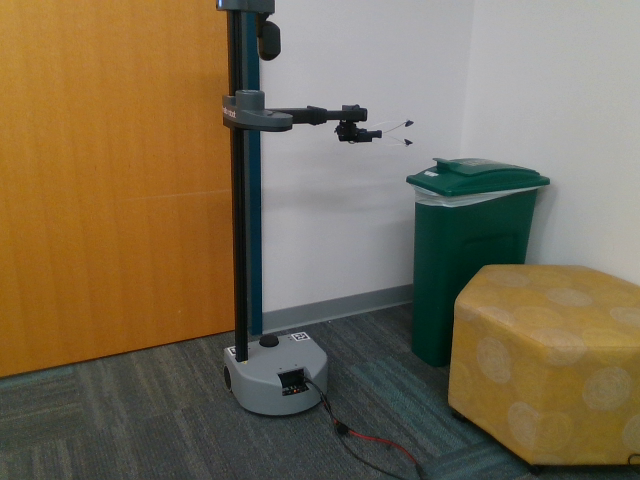}
    \caption{\small Garbage Can}
\end{subfigure}
\begin{subfigure}[b]{0.24\linewidth}
    \includegraphics[width=\linewidth]{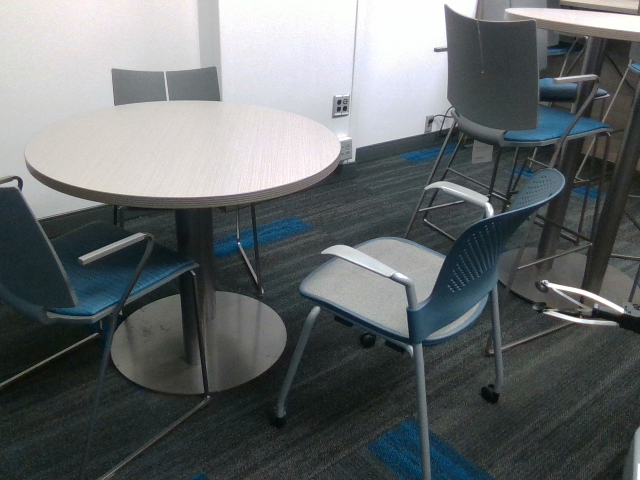}
    \caption{\small Arrange Chair}
\end{subfigure}
\begin{subfigure}[b]{0.24\linewidth}
    \includegraphics[width=\linewidth]{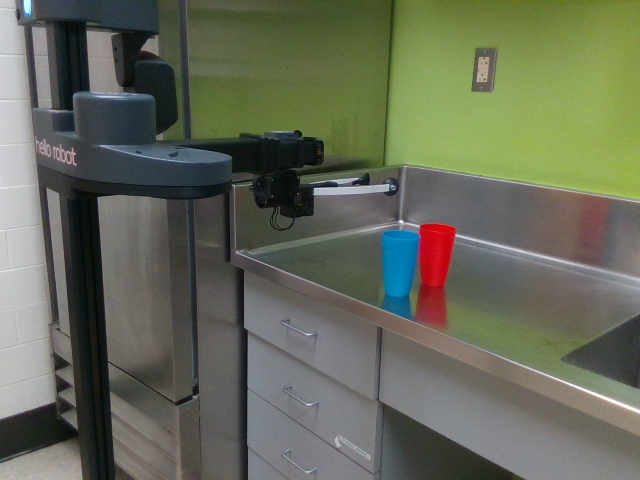}
    \caption{\small Stacking Cups}
\end{subfigure}
\begin{subfigure}[b]{0.24\linewidth}
    \includegraphics[width=\linewidth]{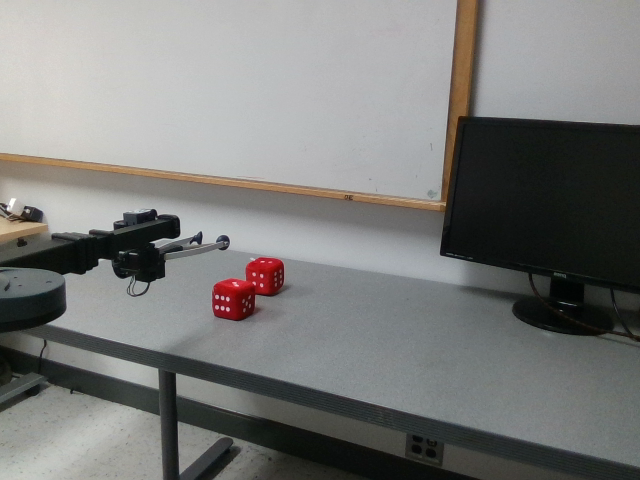}
    \caption{\small Stacking Dice}
\end{subfigure}
\begin{subfigure}[b]{0.24\linewidth}
    \includegraphics[width=\linewidth]{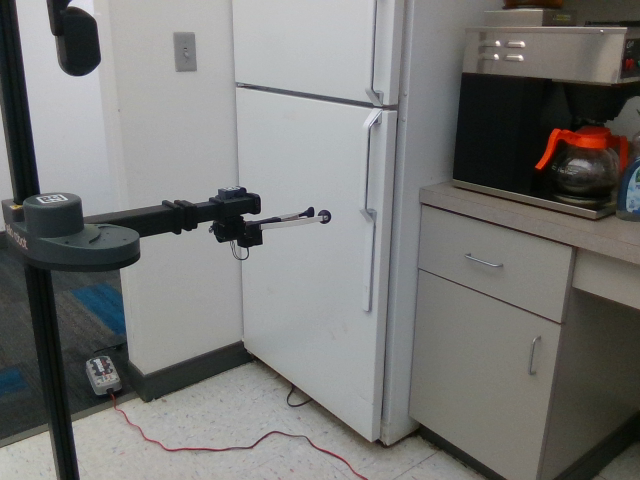}
    \caption{\small Fridge}
\end{subfigure}
\begin{subfigure}[b]{0.24\linewidth}
    \includegraphics[width=\linewidth]{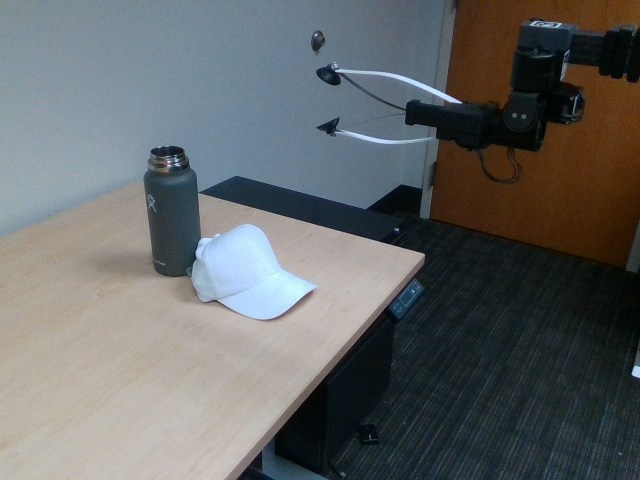}
    \caption{\small Place Hat}
\end{subfigure}
\begin{subfigure}[b]{0.24\linewidth}
    \includegraphics[width=\linewidth]{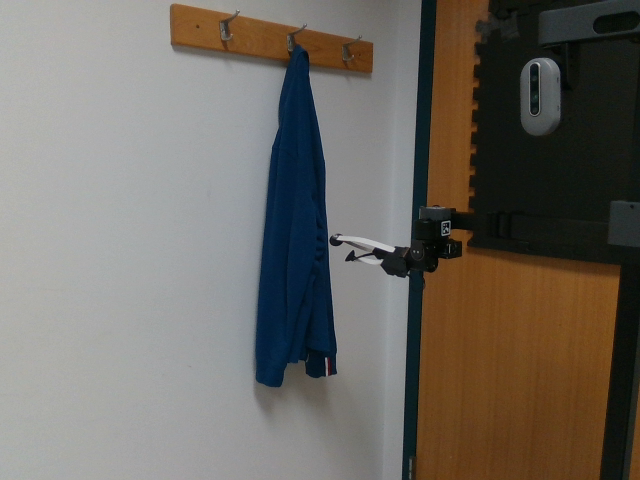}
    \caption{\small Remove Shirt from Hanger}
\end{subfigure}
\begin{subfigure}[b]{0.24\linewidth}
    \includegraphics[width=\linewidth]{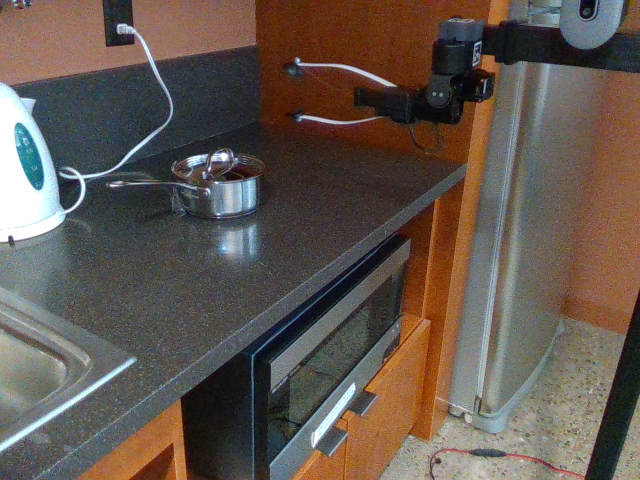}
    \caption{\small Remove Lid}
\end{subfigure}
\begin{subfigure}[b]{0.24\linewidth}
    \includegraphics[width=\linewidth]{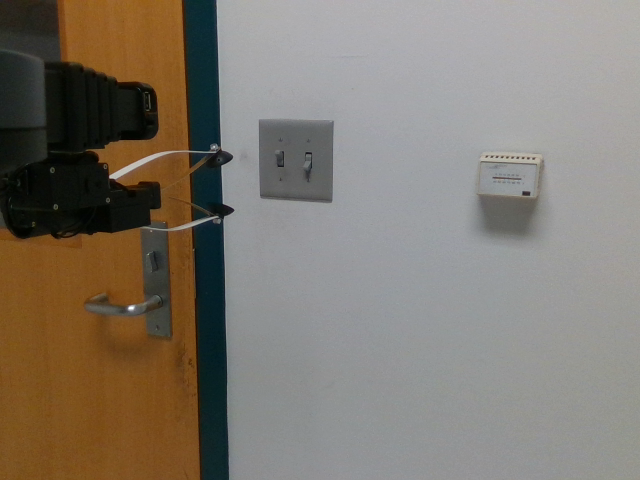}
    \caption{\small Turn Off Light}
\end{subfigure}
\begin{subfigure}[b]{0.24\linewidth}
    \includegraphics[width=\linewidth]{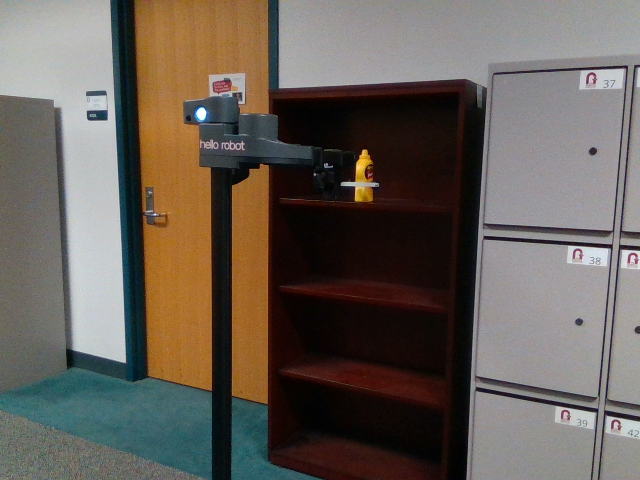}
    \caption{\small Shelf Pick-and-Place}
\end{subfigure}
\begin{subfigure}[b]{0.24\linewidth}
    \includegraphics[width=\linewidth]{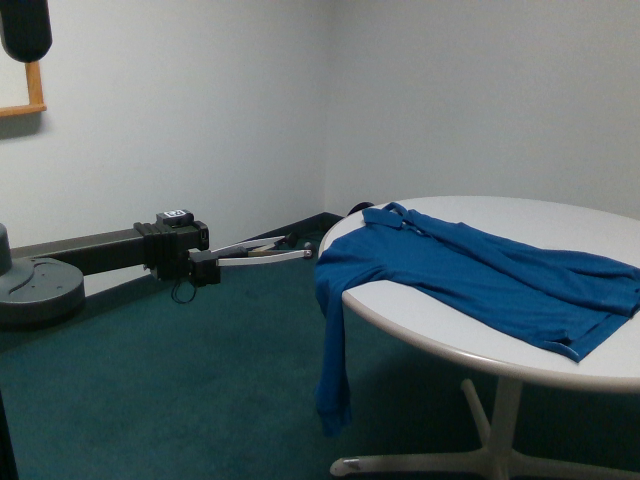}
    \caption{\small Fold Shirt}
\end{subfigure}
\begin{subfigure}[b]{0.24\linewidth}
    \includegraphics[width=\linewidth]{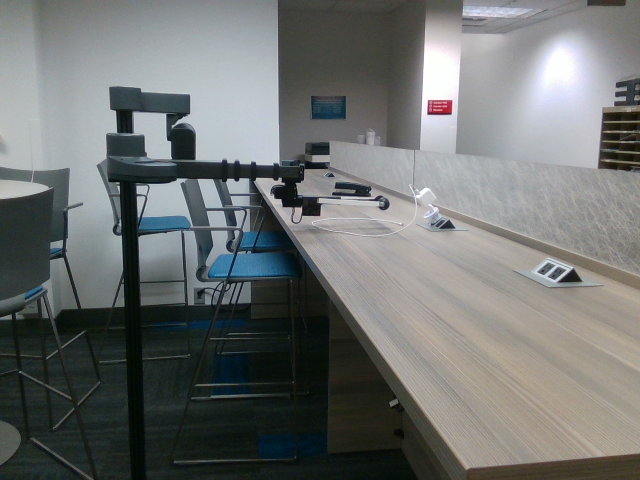}
    \caption{\small Pull Plug from Socket}
\end{subfigure}
\begin{subfigure}[b]{0.24\linewidth}
    \includegraphics[width=\linewidth]{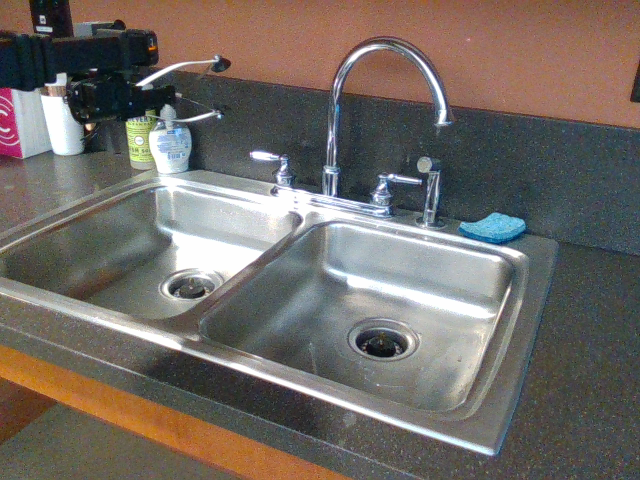}
    \caption{\small Open Tap}
\end{subfigure}
\begin{subfigure}[b]{0.24\linewidth}
    \includegraphics[width=\linewidth]{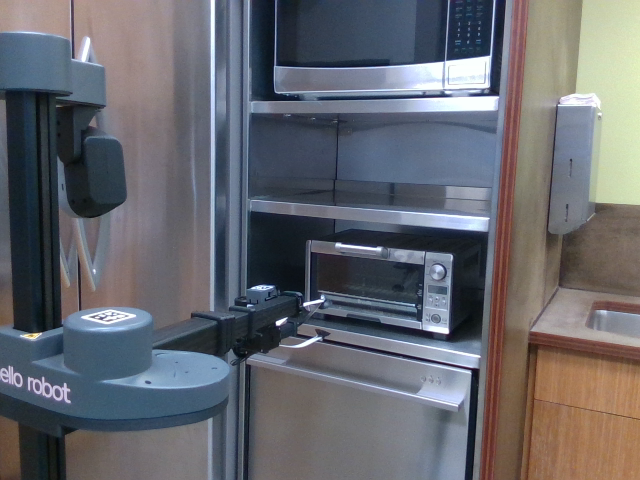}
    \caption{\small Toaster}
\end{subfigure}

\caption{\small Images of our 20 tasks.}
\vspace{-0.1in}
\label{fig:task-list}
\end{figure*}

Our real robot tasks are all in the wild (i.e. outside labs). We perform tasks using everyday objects and in every day locations such as kitchens etc. Due to torque limits on the robots, we had to use non-standard objects for some settings, such as the ball in hoop task, as the robot's gripper cannot grasp a heavier and bigger basketball. In Figure ~\ref{fig:task-list} we present a full list of tasks with images. We show details on train and test objects for our shelf pick-and-place task in Figure~\ref{fig:objects-split}.

\subsubsection{Data Collection}

Human demonstrations are very easy obtain and each takes about 30 seconds to collect. The tasks presented in Figure~\ref{fig:teaser} are trained from one demonstration. During each iteration, 30 samples were taken, and we used the top 10 ranking ones to fit the policy. The Stretch robot \cite{kemp2021design} took less than 1 minute per episode, thus around 20 minutes per iteration.

\subsubsection{Hyper-parameters and Design Choices}

Our policy is a 4 layer MLP, which takes as input an embedding of a demonstration video as well as the prior. The output of the policy is the residual to the prior. We process the human video using an action-recognition pipeline, using features from state-of-the-art pretrained action recognition models such as SlowFast 3D ResNets \cite{feichtenhofer2019slowfast}. We train the policy as a Variational Auto-Encoder, using a KL-divergence loss with weight 0.0005 to train the model and latent dimension = 4. However, larger latent dimensions work well too, but this should be dependent on the size of the action space for the robot (which in our case is 13 dimensional). Our exploration policy is structured in a similar manner. For the human prior, we use the hand-object interaction detector from \citet{100doh}. We employ Copy-Paste Networks \cite{lee2019copy} for inpainting humans and robots from videos. We use action recognition models such as Multi-Moments \cite{monfort2021multi} and SlowFast 3D ResNets \cite{feichtenhofer2019slowfast} as our representation space ($\Phi$) for aligning humand and robot videos. Furthermore, for measuring "change" in the environment we used features from the VGG16  \cite{simonyan2014very} network. Our exploration policy uses the same exact architecture. We use the following video augmentations for our representations: Salt-Pepper Jittering, Random Crops, Gaussian Blurs, Vertical and Horizontal Flips). For the hand-object and wrist detection modules we used default parameters provided by the respective codebases. To smooth the predictions we used the filter from \url{https://docs.scipy.org/doc/scipy/reference/generated/scipy.signal.savgol_filter.html}. We used about 200 labeled images of the robot to train the instance segmentation module, with the default network sizes from the codebase. All of our image and video sizes were 640 x 480. For the policy training module, our optimization approach fits to the top 10 (out of 30) samples.

\subsubsection{Codebases}

We use the following codebases: 
\begin{itemize}
    \item For hand-object detection we use the codebase from \citet{100doh} , \url{https://github.com/ddshan/hand_detector.d2} 
    \item For wrist detection we use FrankMocap \cite{FrankMocap_2021_ICCV}, \url{https://github.com/facebookresearch/frankmocap}
    \item For the instance segmentation module we use code from \url{https://github.com/wkentaro/labelme} to label robot instances, and use code from Detectron2 \cite{wu2019detectron2} (\url{https://github.com/facebookresearch/detectron2}) code provided in \url{https://pytorch.org/tutorials/intermediate/torchvision_tutorial.html} to train an instance segmentation model. 
    \item For the TCN \cite{Sermanet2017TCN} baseline we use code from \url{https://github.com/kekeblom/tcn}
    \item For the CycleGAN baseline we use code from \url{https://github.com/Lornatang/CycleGAN-PyTorch} 
    \item Our Inpainting model \cite{lee2019copy}: \url{https://github.com/shleecs/Copy-and-Paste-Networks-for-Deep-Video-Inpainting} 
    \item We use the model from \citet{monfort2021multi} (\url{https://github.com/zhoubolei/moments_models}) to compute representations for our cost functions
    \item Our offline RL baselines are from \citet{seno2021d3rlpy} (\url{https://github.com/takuseno/d3rlpy})
    \item \url{https://github.com/okankop/vidaug}
    
\end{itemize}

\end{document}